\documentclass{article}
\usepackage{authblk}

\usepackage[utf8]{inputenc}
\usepackage{amsmath}
\usepackage{amssymb}
\usepackage{amsthm}
\usepackage{xcolor}
\usepackage{mathtools}
\usepackage{soul}
\usepackage{url}  
\usepackage{algorithm}
\usepackage{tcolorbox}
\usepackage[noend]{algpseudocode}
\usepackage{subcaption}
\usepackage{hyperref}
\usepackage{enumitem}
\usepackage{booktabs} 
\usepackage{multirow} 
\usepackage{siunitx} 
\usepackage{amsfonts} 
 
\usepackage{multirow}
\usepackage{todonotes}

\usepackage{todonotes}
\usepackage{booktabs}
\usepackage{cleveref}
\usepackage{soul}
\usepackage{natbib}
\usepackage{comment}

\newcommand{\new}[1]{\textcolor{black}{#1}}
\usepackage{xspace}
\makeatletter
\DeclareRobustCommand\onedot{\futurelet\@let@token\@onedot}
\def\@onedot{\ifx\@let@token.\else.\null\fi\xspace}

\def\ie{\emph{i.e.~}}
\def\eg{\emph{e.g.~}}
\makeatother

\newtheorem{lemma}{Lemma}
\newtheorem{theorem}{Theorem}
\newtheorem{corollary}{Corollary}
\newtheorem{definition}{Definition}

\begin{document}

\title{\bf Bounding Membership Inference}

\date{}

\author[]{Anvith Thudi}
\author[]{Ilia Shumailov}
\author[]{Franziska Boenisch}
\author[]{Nicolas Papernot}
\affil[]{University of Toronto and Vector Institute}

\maketitle

\begin{abstract}
    Differential Privacy (DP) is the de facto standard for reasoning about the privacy guarantees of a training algorithm. Despite the empirical observation that DP reduces the vulnerability of models to existing membership inference (MI) attacks, a theoretical underpinning as to why this is the case is largely missing in the literature. In practice, this means that models need to be trained with DP guarantees that greatly decrease their accuracy. 
    
    In this paper, we provide a tighter bound on the positive accuracy (\ie attack precision) of any MI adversary when a training algorithm provides $(\varepsilon, \delta)$-DP. Our bound informs the design of a novel privacy amplification scheme: an effective training set is sub-sampled from a larger set prior to the beginning of training. We find this greatly reduces the bound on MI positive accuracy. As a result, our scheme allows the use of looser DP guarantees to limit the success of any MI adversary; this ensures that the model's accuracy is less impacted by the privacy guarantee. While this clearly benefits entities working with far more data than they need to train on, it can also improve the accuracy-privacy trade-off on benchmarks studied in the academic literature. Consequently, we also find that subsampling decreases the effectiveness of a state-of-the-art MI attack (LiRA) much more effectively than training with stronger DP guarantees on MNIST and CIFAR10. We conclude by discussing implications of our MI bound on the field of machine unlearning.

\end{abstract}

\section{Introduction}

Differential Privacy (DP) \citep{dwork2006calibrating} is employed extensively to reason about privacy guarantees in a variety of settings \citep{dwork2008differential}. Recently, DP started being used to give privacy guarantees for the training data of deep neural networks (DNNs) learned through stochastic gradient descent (SGD) \citep{abadi2016deep}. However, even though DP provides privacy guarantees and bounds the worst case privacy leakage, it is not immediately clear how these guarantees bound the accuracy of known existing forms of privacy infringement attacks.

At the time of writing, the most practical attack on the privacy of DNNs is Membership Inference (MI) \citep{shokri2017membership}, where an adversary predicts whether or not a model used a particular data point for training; note that this is quite similar to the hypothetical adversary at the core of the game instantiated in the definition of DP. 
MI attacks saw a strong interest by the community, and several improvements and renditions were proposed since its inception \citep{sablayrolles2019white,choquette2021label, maini2021dataset, hu2021membership}. Having privacy guarantees would desirably defend against MI, and in fact, current literature highlights that DP does give an upper bound on the performance of MI adversaries~\citep{yeom2018privacy,erlingsson2019we,sablayrolles2019white,jayaraman2020revisiting}. 

In this paper, we first propose a tight bound on MI performance for training algorithms that provide $\varepsilon$-DP; in particular, the quantity of interest for our bound is MI positive accuracy (\ie the MI attack's precision). We focus on positive accuracy since predicting incorrectly often (as measured by accuracy) is not an effective attack strategy in practice, whereas being always correct that a given datapoint appears in a training set is advantageous. Bounding positive accuracy, as is done in this paper, removes this worst-case scenario by bounding the certainty any adversary can have when predicting positively. The key insight for our bound is that an MI adversary only ever has a finite set of points they suspect were used during training. This allows us to apply a counting lemma to simplify terms in the equation for MI positive accuracy. Our bound is tight up to the looseness of the DP guarantee holding across all datasets. We also extend our result to bound MI positive accuracy for $(\varepsilon,\delta)$-DP, a common relaxation of $\varepsilon$-DP.

Perhaps most importantly, in obtaining our bound, we expose how MI positive accuracy is influenced by a form of privacy amplification where the training dataset itself is sub-sampled from a larger dataset. Amplification is a technique pervasively found in work improving the analysis of DP learners like DP-SGD~\citep{abadi2016deep}. It allows one to prove tighter bounds on privacy leakage. Here, we observe the effect of our amplification on lowering MI positive accuracy is significantly stronger than the effect of previous forms of amplifications used to obtain tigther $\varepsilon$-DP guarantees. 

In particular, amplification from subsampling achieves a greater effect on MI positive accuracy than batch sampling -- a common privacy amplification scheme for training deep neural networks. This finding has immediate consequences for entities with far more data than they need to train on: subsampling data before training will greatly contribute to reducing the risk of membership inference. We also observe benefits for bechmarks commonly studied in the academic literature like MNIST and CIFAR10. For instance, we find that the success of a state-of-the-art MI attack (LiRA~\cite{carlini2022membership}) experiences an outweighted effect from subsampling compared to reducing the $\varepsilon$ in the DP guarantee. This confirms that studying MI positive accuracy exposes factors to control MI performance that allow training with higher accuracy than simply reasoning with the DP guarantee would allow.

Our bound also has consequences for the problem of unlearning (or data forgetting in ML) introduced by~\citet{cao2015towards}. In particular the MI accuracy on the point to be unlearned is a popular measure for how well a model has unlearned it  \citep{baumhauer2020machine,graves2020amnesiac,golatkar2020forgetting,golatkar2020eternal}. 
However empirical verification of the MI accuracy can be open-ended, as it is subjective to the attack employed. This can lead one to underestimate the need for unlearning by overestimating how well a model has already unlearned a point. Theoretical bounds on all MI attacks, such as the one proposed in this work, circumvent this issue; a bound on the precision of MI attacks, that is, the probability that a data point was used in the training dataset, indicates a limitation for any entity to discern if the model has been trained on that data point. In the case when this probability is sufficiently low (where sufficiently is defined apriori), one can then claim to have unlearned by achieving a model sufficiently likely to have not come from training with the data point. Our analysis shows that, if dataset sub-sampling is used, one can unlearn under this definition by training with a relatively large $\varepsilon$-DP (and thus have less cost to accuracy performance than previously observed). 

To summarize, our contributions are:
\begin{itemize}
    \item We present a bound on MI \new{positive accuracy (\ie attack precision)} for $\varepsilon$-DP and $(\varepsilon,\delta)$-DP. This bound is tight up to the looseness of the DP guarantee holding across all datasets. 
    \item We further demonstrate how to lower this bound using a novel privacy amplification scheme built on dataset subsampling; we explain the theoretical benefits of this for entities that do not need to train on all their data.
    \item We show this new factor our bound highlights also has empirical benefits on CIFAR10 and MNIST: dataset subsampling has an outweighed effect on the success of a state-of-the-art MI attack (LiRA).
    \item We discuss the benefits of such bounds to machine unlearning as a rigorous way to use MI as a metric for unlearning.
\end{itemize}

\section{Background}

\subsection{Differential Privacy}

Differential privacy (DP) \citep{dwork2006calibrating} bounds how different the outputs of a function on adjacent inputs can be in order to provide privacy guarantees for the inputs. More formally, a function $F$ is $\varepsilon$-DP if for all adjacent inputs $x$ and $x'$ (\ie inputs with Hamming distance of 1) we have for all sets $S$ in the output space:
\begin{equation}
    \mathbb{P}(F(x) \in S) \leq e^{\varepsilon}\mathbb{P}(F(x') \in S).
\end{equation}
There also is the more relaxed notion of $(\varepsilon,\delta)$-DP which is defined for the same setup as above, but introduces a parameter $\delta \in (0,1]$, such that $\mathbb{P}(F(x) \in S) \leq e^{\varepsilon}\mathbb{P}(F(x') \in S) + \delta$. Notably, $(\varepsilon,\delta)$-DP is used for functions where it is more natural to work with $\ell_2$ metrics on the input space, having to do with how DP guarantees are obtained.

To achieve DP guarantees, one usually introduces noise to the output of the function $F$.
The amount of noise is calibrated to the maximal $\ell_2$ or $\ell_1$ difference between all possible outputs of the function on adjacent datasets (also called \emph{sensitivity}). Significant progress was achieved on minimizing the amount of noise needed for a given sensitivity \citep{balle2018improving}, and on how DP guarantees scale when composing multiple DP functions \citep{dwork2010boosting,kairouz2015composition}.

\citet{abadi2016deep,song2013stochastic,bassily2014private} demonstrated a method to make the final model returned by mini-batch SGD $(\varepsilon,\delta)$-DP with respect to its training dataset by bounding the sensitivity of gradient updates during mini-batch SGD and introducing Gaussian noise to each update. This approach became the de-facto standard for DP guarantees in DNNs. However, the adoption is still greatly limited because of an observed trade-off between privacy guarantees and model utility. At the time of writing there is still no feasible way to learn with low $\varepsilon$ and high accuracy; what is more, past work~\citep{jagielski2020auditing} observed a gap between observed and claimed privacy, which suggests that DP-analysis, so far, may be too loose.

However, more recently \citet{nasr2021adversary} showed (using statistical tests and stronger MI adversaries) that the current state of the art methods of achieving $(\varepsilon,\delta)$-DP guarantees for deep learning are tight, in contrast to the gap \citet{jagielski2020auditing} observed. This suggests that there is not much more improvement to be gained by studying how to improve the $(\varepsilon,\delta)$-DP guarantee from a given amount of noise or improving composition rules. Facing this, future improvements in DP training would lie in understanding the guarantees that DP provides against the performance of relevant attacks. Once more informed about the guarantees required during training,
this enables the use of looser guarantees if one is only interested in defending against a specific set of attacks.%
\footnote{It is worth noting that \cite{nasr2021adversary} showed that current analytic upper bounds on DP guarantees are tight, measuring them empirically with various strong privacy adversaries. Although results do suggest that bounds match, the paper did not investigate how DP guarantees limit performance of the adversary.}

\subsection{Membership Inference}

\citet{shokri2017membership} introduced an MI attack against DNNs, which leveraged shadow models (models with the same architecture as the target model) trained on similar data in order to train a classifier which, given the outputs of a model on a data point, predicts if the model was trained on that point or not. Since the introduction of this initial attack, the community has proposed several improved and variations of the original MI attack \citep{ yeom2018privacy, salem2018ml,sablayrolles2019white,truex2019demystifying, jayaraman2020revisiting, maini2021dataset,choquette2021label}.

As mentioned previously, \citet{nasr2021adversary} demonstrated that current analysis of DP-SGD can not be significantly improved, and so further privacy improvements will be in understanding how DP bounds privacy attacks. MI attacks are currently the main practical threat to the privacy of the data used to train DNNs. Hence, tighter bounds on MI attacks when using DP would help entities reason about what DP parameters to train with; this is particularly relevant when stronger DP guarantees are known to also reduce performance and fairness \citep{bagdasaryan2019differential}.

In our paper, we will work with the following (abstract) notion of MI in giving our bounds. \new{In particular, we consider MI as a decision problem over the space of models (\ie weights) such that if the model is in a particular set $S$, it predicts $\mathbf{x}^*$ is in the training dataset, else not. That is, to every adversary we associate a set $S$. We then have the following definitions of its "positive accuracy", \ie accuracy when predicting $\mathbf{x}^*$ is in the training set (also called precision or positive predictive value):}

\begin{definition}[MI accuracy]
\label{def:MI_acc}
\new{The positive accuracy of a MI adversary, associated to the set $S$, is $\mathbb{P}(\mathbf{x}^* \in D|S)$. Similarly, the negative accuracy is $\mathbb{P}(\mathbf{x}^* \notin D|S) = 1 - \mathbb{P}(\mathbf{x}^* \in D|S)$,}

\end{definition}
where $D$ is the training dataset and $\mathbb{P}(\mathbf{x}^* \in D|S)$ is the probability a data point $\mathbf{x}^*$ was in the training dataset used to obtain the models in the set $S$ (\ie this is a probability over datasets). We explain more about where the randomness is introduced (in particular the probability involved in obtaining a training dataset) in Section~\ref{ssec:scenario}.

\subsection{Previous Bounds}
\label{ssec:Prev_bounds}

Before describing past bounds on MI accuracy, we have to formally define the attack setting past work considered. Two of the main bounds \citep{yeom2018privacy,erlingsson2019we} focused on an experimental setup first introduced by \citet{yeom2018privacy}. In this setting, an adversary $f$ is given a datapoint $\mathbf{x}^*$ that is $50\%$ likely to have been used to train a model $S$ or not. The adversary then either predicts $1$ if they think it was used, or $0$ otherwise. Let $b = 1~\text{or}~0$ indicate if the datapoint was or was not used for training, respectively;
we say the adversary was correct if their prediction matches $b$. We then define the adversary's advantage as improvement in accuracy over the $50\%$ baseline of random guessing, or more specifically $2(A(f)-0.5)$ where $A(f)$ is the accuracy of~$f$.

For such an adversary operating in a setting where data is equally likely to be included or not in the training dataset, \citet{yeom2018privacy} showed that they could bound the advantage of the adversary by $e^{\varepsilon} - 1$ when training with $\varepsilon$-DP. In other words, they showed that they could bound the accuracy of the MI adversary by $e^{\varepsilon}/2$. Their proof used the fact that the true positive rate~(TPR) and false positive rate~(FPR) of their adversary could be represented as expectations over the different data points in a dataset, and from that introduced the DP condition to obtain their MI bound, noting that MI advantage is equivalent to TPR - FPR.

\citet{erlingsson2019we} improved on the bound developed by \citet{yeom2018privacy} for an adversary operating under the same condition by utilizing a proposition given by \citet{hall2013differential} on the relation between TPR and FPR for an $(\varepsilon,\delta)$-DP function. Based on these insights, \citet{erlingsson2019we} bounded the membership advantage by $1-e^{-\varepsilon} + \delta e^{-\varepsilon}$, which is equivalent to bounding the accuracy of the adversary by $1-e^{-\varepsilon}/2$ when $\delta = 0$ (\ie in $\varepsilon$-DP). \new{Finally, improving on \citet{erlingsson2019we} by using propositions given by \citet{kairouz2015composition}, \citet{humphries2020differentially} bounded the membership advantage by $\frac{e^{\varepsilon}-1+2\delta}{e^{\varepsilon}+1}$. This translates to an accuracy bound of $\frac{2e^{\varepsilon}+2\delta}{2(e^{\varepsilon}+1)}$ which is the tightest MI accuracy bound we are aware of.} We will see that as a corollary of our results we can match \citet{humphries2020differentially} bound on accuracy, but moreover, our proof shows this bound is tight upto the looseness of the DP guarantee holding over all datasets.

Other works, similar to our setup in~\Cref{ssec:scenario}, considered a more general setting where the probability of sampling a datapoint in the dataset can vary. For $\varepsilon$-DP, \citet{sablayrolles2019white} bounded the probability of a datapoint $\mathbf{x}^*$ being used in the training set of a model (\ie, the accuracy of an attacker who predicted the datapoint was in the dataset of the model \new{which we refer to as positive accuracy}) by $\mathbb{P}_{\mathbf{x}^*}(1) + \frac{\varepsilon}{4}$ where $\mathbb{P}_{\mathbf{x}^*}(1)$ is the probability of the datapoint being in the dataset. \new{Recent concurrent work by \citet{mahloujifar2022optimal} improves the bound on positive accuracy (also called precision or positive predictive value) to $\frac{1}{1+e^{-\varepsilon}}$ in the case of $\mathbb{P}_{\mathbf{x}^*}(1) = 0.5 $ and all other data is fixed.} Our bound on positive accuracy both improves over \citet{sablayrolles2019white} and generalizes to a more general setting than the one  \citet{mahloujifar2022optimal} is limited to.

Finally, \citet{jayaraman2020revisiting} bounded the positive predictive value of an attacker on a model trained with $(\varepsilon,\delta)$-DP when the FPR is fixed. 
Similarly, Jayaraman et al. further bounded MI advantage under the experiment described by \citet{yeom2018privacy} for a fixed FPR. Note, that both our work and the previously mentioned bounds are independent of FPR. In particular \citet{erlingsson2019we} followed a similar technique to \citet{jayaraman2020revisiting}, but were able to drop the FPR term using a proposition relating TPR to FPR \citep{hall2013differential}.

\section{Bounds on Positive and Negative Accuracy of Membership Inference }

\subsection{The Setting}
\label{ssec:scenario}

Our setting is more general than the one introduced by \citet{yeom2018privacy} and, as we note later, a specific instance of it can be reduced to their setting. In particular, we formalize how an entity samples data into the training dataset, and proceed with our analysis from there.

The intuition for our formalization is the existence of some finite data superset containing all the data points that an entity could have in their training dataset.
Yet, any one of these datapoints only has some probability of being sampled into the training dataset. For example, this larger dataset could consist of all the users that gave an entity access to their data, and the probability comes from the entity randomly sampling the data to use in their training dataset. 
This randomness can be a black-box such that not even the entity knows what data was used to train.
In essence, this is the setting \citet{jayaraman2020revisiting} considers, though in their case, the larger dataset is implicit and takes the form of an arbitrary distribution. We can then imagine that the adversary (or perhaps an arbitrator in an unlearning setup) knows the larger dataset and tries to infer whether a particular point was used in the training dataset. The particular general MI attack we analyze and bound is based on this setting.

Specifically, let the individual training datasets $D$ be constructed by sampling from a finite countable set where all datapoints are unique and sampled independently, \ie from some larger set $\{\mathbf{x}_1,\cdots,\mathbf{x}_N\}$.
That is if $D = \{\mathbf{x}_1,\mathbf{x}_2,\cdots,\mathbf{x}_n\}$ then the probability of sampling $D$ is $\mathbb{P}(D) = \mathbb{P}_{\mathbf{x}_1}(1) \mathbb{P}_{\mathbf{x}_2}(1) \cdots \\ \mathbb{P}_{\mathbf{x}_n}(1)\mathbb{P}_{\mathbf{x}_{n+1}}(0)\cdots\mathbb{P}_{\mathbf{x}_N}(0)$, where $\mathbb{P}_{\mathbf{x}_i}(1)$ is probability of drawing $\mathbf{x}_i$ into the dataset and $\mathbb{P}_{\mathbf{x}_i}(0)$ is the probability of not.

We define $\mathfrak{D}$ as the set of all datasets. Let now $\mathfrak{D_{\mathbf{x}^*}}$ be the set of all datasets that contain a particular point $\mathbf{x}^* \in \{\mathbf{x}_1,\cdots,\mathbf{x}_N\}$, that is $\mathfrak{D_{\mathbf{x}^*}} = \{D~:~\mathbf{x}^* \in D\}$. Similarly let $\mathfrak{D_{\mathbf{x}^*}}'$ be the set of all datasets that do not contain $\mathbf{x}^*$, \ie $\mathfrak{D_{\mathbf{x}^*}}' = \{D'~:~\mathbf{x}^* \notin D'\}$. Note $\mathfrak{D} = \mathfrak{D_{\mathbf{x}^*}} \cup  \mathfrak{D_{\mathbf{x}^*}}'$ by the simple logic that any dataset has or does not have $\mathbf{x}^*$ in it. We then have the following lemma.

\begin{lemma}
\label{lem:bijective}
$\mathfrak{D_{\mathbf{x}^*}}$ and $\mathfrak{D_{\mathbf{x}^*}}'$ are in bijective correspondence with $\mathbb{P}(D)\frac{\mathbb{P}_{\mathbf{x}^*}(0)}{\mathbb{P}_{\mathbf{x}^*}(1)} = \mathbb{P}(D')$ for $D \in \mathfrak{D_{\mathbf{x}^*}}$ and $D' \in \mathfrak{D_{\mathbf{x}^*}}'$ that map to each other under the bijective correspondence.
\end{lemma}

\begin{proof}
Note that for a given $D=\{\mathbf{x}_1,\cdots,\mathbf{x}_n\} \in \mathfrak{D_{\mathbf{x}^*}}$, $D' = D/\mathbf{x}^* \in \mathfrak{D_{\mathbf{x}^*}}'$ is unique (\ie the map by removing $\mathbf{x}^*$ is injective) and similarly for a given $D' \in \mathfrak{D_{\mathbf{x}^*}}'$ $D = D' \cup \mathbf{x}^* $ is unique (\ie the map by adding $\mathbf{x}^*$ is injective). Thus, we have injective maps running both ways which are the inverses of each other.
As a consequence, we have $\mathfrak{D_{\mathbf{x}^*}}$ and $\mathfrak{D_{\mathbf{x}^*}}'$ are in bijective correspondence. 

Now if the larger set of datapoints is $\{\mathbf{x}_1\cdots\mathbf{x}_{n-1},\mathbf{x}^*,\mathbf{x}_{n}\cdots\mathbf{x}_{N}\}$ letting $D = \{\mathbf{x}_1\cdots\mathbf{x}_{n-1}\}\cup \mathbf{x}^*$ and $D' = \{\mathbf{x}_1\cdots\mathbf{x}_{n-1}\}$ be any pair of datasets that map to each other by the above bijective map, then note $\mathbb{P}(D) = \mathbb{P}_{\mathbf{x}_1}(1)\mathbb{P}_{\mathbf{x}_2}(1) \cdots \\ \mathbb{P}_{\mathbf{x}_{n-1}}(1)\mathbb{P}_{\mathbf{x}^*}(1)\cdots\mathbb{P}_{\mathbf{x}_{n+1}}(0)\cdots\mathbb{P}_{\mathbf{x}_N}(0)$ and $\mathbb{P}(D') = \mathbb{P}_{\mathbf{x}_1}(1)\mathbb{P}_{\mathbf{x}_2}(1)  \cdots \\ \mathbb{P}_{\mathbf{x}_{n-1}}(1)\mathbb{P}_{\mathbf{x}^*}(0)\cdots\mathbb{P}_{\mathbf{x}_{n+1}}(0)\cdots\mathbb{P}_{\mathbf{x}_N}(0)$. In particular we have $\mathbb{P}(D)  \frac{\mathbb{P}_{\mathbf{x}^*}(0)}{\mathbb{P}_{\mathbf{x}^*}(1)}  = \mathbb{P}(D')$.

\end{proof}

Once some dataset $D$ is obtained, we call $H$ the training function which takes in $D$ and outputs a model $M$ as a set of weights in the form of a real vector. Recall that $H$ is $\varepsilon$-DP if for all adjacent datasets $D$ and $D'$ and any set of model(s) $S$ in the output space of $H$ (\ie some weights) we have: $\mathbb{P}(H(D) \in S) \leq e^{\varepsilon}\mathbb{P}(H(D') \in S)$. It should be noted that from now on we assume that the set $S$ has a non-zero probability to be produced by $H$. This is sensible as we are not interested in MI attacks on sets of models that have $0$ probability to come from training; note also if $\mathbb{P}(H(D) \in S) = 0$, then $\mathbb{P}(H(D') \in S) = 0$ for all adjacent $D'$ as $ 0 \leq \mathbb{P}(H(D') \in S) \leq e^{\varepsilon}\mathbb{P}(H(D) \in S) = 0$, and thus the probability is $0$ for all countable datasets as we can construct any dataset by removing and adding a data point (which does not change the probability if it is initially $0$) countably many times.

\subsection{Our Bounds}
\label{ssec:main_result}

We now proceed to use Lemma \ref{lem:bijective} to bound the positive and negative accuracy of MI, as stated in Definition \ref{def:MI_acc}, for a training function $H$ that is $\varepsilon$-DP under the data-sampling setting defined earlier. Our approach differs from those we discussed in~\Cref{ssec:Prev_bounds} in that we now focus on the definition of the conditional probability $\mathbb{P}(\mathbf{x}^* \in D|S)$ as a quotient; finding a bound then reduces to finding a way to simplify the quotient with the $\varepsilon$-DP definition, which we achieve using Lemma~\ref{lem:bijective}.

What follows are the bounds, with \Cref{sec: sampling_effect},\ref{ssec:data_deletion}, and ~\ref{sec:discussion} expanding on the consequences of the bounds.

\begin{theorem}[DP bounds MI positive accuracy]
\label{thm:MI_pos_acc}

\new{For any MI attack on datapoint $\mathbf{x}^*$ (\ie with any associated set $S$)}, and the training process $H$ is DP with $\varepsilon$, its \new{positive} accuracy is upper-bounded by $\left({1+\frac{e^{-\varepsilon}\mathbb{P}_{\mathbf{x}^*}(0)}{\mathbb{P}_{\mathbf{x}^*}(1)}}\right)^{-1}$ and lower bounded by $\left({1+\frac{e^{\varepsilon}\mathbb{P}_{\mathbf{x}^*}(0)}{\mathbb{P}_{\mathbf{x}^*}(1)}}\right)^{-1}$, where $\mathbb{P}_{\mathbf{x}^*}(1)$ is the probability of drawing $\mathbf{x}^*$ into the dataset \footnote{\new{We can obtain a bound against all points by using the largest $\mathbb{P}_{\mathbf{x}}(1)$.}}.

\end{theorem}

The high level idea for the proof is to rewrite positive accuracy as a fraction, and use Lemma 1 (applicable since we only deal with countable sets) to simplify terms alongside the $\varepsilon$-DP condition. The $\varepsilon$-DP gives inequalities in both directions, providing us with the lower- and upper-bounds \footnote{\new{Note that the previous bounds apply to any adversary. However if one wants to give a higher "existence lower-bound", i.e an instantiated adversary with a given accuracy, there is the trivial adversary which always predicts the datapoint was used to train and obtains positive accuracy $\mathbb{P}_{\mathbf{x}^*}(1)$.}}.

\begin{proof}

The positive accuracy of an adversary $f$ \new{associated to any set $S$ is}:

\begin{equation}
    \mathbb{P}(\mathbf{x}^*|S) = \frac{\sum_{D \in \mathfrak{D}_{\mathbf{x}^*}}\mathbb{P}(H(D) \in S)\mathbb{P}(D)}{\sum_{D \in \mathfrak{D}}\mathbb{P}(H(D) \in S)\mathbb{P}(D)}.
    \label{eq:accuracy}
\end{equation}

By the observation $\mathfrak{D} = \mathfrak{D_{\mathbf{x}^*}} \cup  \mathfrak{D_{\mathbf{x}^*}}'$ we have that the denominator can be split into $\sum_{D \in \mathfrak{D}_{\mathbf{x}^*}}\mathbb{P}(H(D) \in S)\mathbb{P}(D) + \sum_{D' \in \mathfrak{D}_{\mathbf{x}^*}'}\mathbb{P}(H(D') \in S)\mathbb{P}(D')$. 

By Lemma \ref{lem:bijective}, we can replace the $D' \in \mathfrak{D}_{\mathbf{x}^*}'$ in the second sum by $D \in \mathfrak{D}_{\mathbf{x}^*}$ and replace $\mathbb{P}(D')$ by $\mathbb{P}(D)\frac{\mathbb{P}_{\mathbf{x}^*}(0)}{\mathbb{P}_{\mathbf{x}^*}(1)}$. For $\mathbb{P}(H(D') \in S)$ note by $H$ being $\varepsilon$-DP we have $\mathbb{P}(H(D') \in S) \geq e^{-\varepsilon}\mathbb{P}(H(D) \in S)$ and so with the previous replacements we have that the denominator is greater than $(1+\frac{e^{-\varepsilon}\mathbb{P}_{\mathbf{x}^*}(0)}{\mathbb{P}_{\mathbf{x}^*}(1)})\cdot\sum_{D \in \mathfrak{D_{\mathbf{x}^*}}}\mathbb{P}(H(D) \in S)\mathbb{P}(D)$.

Thus, the \new{positive} accuracy of $f$ is $\leq \frac{1}{1+\frac{e^{-\varepsilon}\mathbb{P}_{\mathbf{x}^*}(0)}{\mathbb{P}_{\mathbf{x}^*}(1)}}$ (\ie the upper bound). If instead we used the fact that $\mathbb{P}(H(D') \in S) \leq e^{\varepsilon}\mathbb{P}(H(D) \in S)$, we would find that the \new{positive} accuracy of $f$ is $\geq \frac{1}{1+\frac{e^{\varepsilon}\mathbb{P}_{x^*}(0)}{\mathbb{P}_{\mathbf{x}^*}(1)}}$ (\ie the lower bound).

\end{proof}

By the definition of negative accuracy, we have the following corollary:

\begin{corollary}[DP bounds MI negative accuracy]
\label{cor:MI_neg_acc}
\new{For any MI attack on datapoint $\mathbf{x}^*$ (\ie with any associated set $S$)}, and the training process $H$ is DP with $\varepsilon$, then the \new{negative} accuracy is upper-bounded by $\left({1+\frac{e^{-\varepsilon}\mathbb{P}_{\mathbf{x}^*}(1)}{\mathbb{P}_{\mathbf{x}^*}(0)}}\right)^{-1}$ and lower-bounded by $\left({1+\frac{e^{\varepsilon}\mathbb{P}_{\mathbf{x}^*}(1)}{\mathbb{P}_{\mathbf{x}^*}(0)}}\right)^{-1}$, where $\mathbb{P}_{\mathbf{x}^*}(1)$ is the probability of drawing $\mathbf{x}^*$ into the dataset.
\end{corollary}

\begin{proof}

Immediately follows from Theorem \ref{thm:MI_pos_acc} and Defintion \ref{def:MI_acc}, as if $\mathbb{P}(\mathbf{x}^*|S) \geq \frac{1}{1+\frac{e^{\varepsilon}\mathbb{P}_{x^*}(0)}{\mathbb{P}_{x^*}(1)}}$ then $$1- \mathbb{P}(\mathbf{x}^*|S) \leq 1 - \frac{1}{1+\frac{e^{\varepsilon}\mathbb{P}_{x^*}(0)}{\mathbb{P}_{x^*}(1)}} = \frac{1}{1+\frac{e^{-\varepsilon}\mathbb{P}_{x^*}(1)}{\mathbb{P}_{x^*}(0)}}.$$.

Similarly, we get $1- \mathbb{P}(\mathbf{x}^*|S) \geq 1 - \frac{1}{1+\frac{e^{-\varepsilon}\mathbb{P}_{x^*}(0)}{\mathbb{P}_{x^*}(1)}} = \frac{1}{1+\frac{e^{\varepsilon}\mathbb{P}_{x^*}(1)}{\mathbb{P}_{x^*}(0)}}$.

\end{proof}

Note that in the case $\mathbb{P}_{\mathbf{x}^*}(1) = \mathbb{P}_{\mathbf{x}^*}(0) = 0.5$, the bounds given by Theorem \ref{thm:MI_pos_acc} and Corollary \ref{cor:MI_neg_acc} are identical.
Therefore, as $f(\mathbf{x}^*,S)$ must output either $0$ or $1$, we have a more general claim that the attack accuracy (maximum of positive or negative accuracy) is always bounded by the same values given by Theorem~\ref{thm:MI_pos_acc}.

\paragraph{Remark on tightness} Note that the bounds in Theorem~\ref{thm: eps_delta_bound} are tight if all the DP inequalities we use (specifically $\mathbb{P}(H(D') \in S) \geq e^{-\varepsilon}\mathbb{P}(H(D) \in S)~\forall D\in \mathfrak{D}_{\mathbf{x}^*}$) are tight.
This is due to those DP inequalities being the only inequalities appearing in the proof. 
Hence, any future improvement would have to go beyond the DP-condition and incorporate the looseness of the DP guarantee on different datasets.
This means that our bound is optimal when no such knowledge is provided. 
Recent concurrent work by \cite{mahloujifar2022optimal} does just that by studying MI accuracy bounds given by a privacy mechanism, beyond just the $(\varepsilon,\delta)$-DP guarantees it yields.
We remark now that MI accuracy, which is what \cite{mahloujifar2022optimal} studies, and MI positive accuracy (what we study) have divergent analyses in the case of $(\varepsilon,\delta)$-DP, which will be seen in the following section (\S~\ref{sec:eps_delta}).

\section{Limitations of $(\varepsilon,\delta)-$DP}
\label{sec:eps_delta}

We now illustrate the shortcomings of $(\varepsilon,\delta)$-DP in giving similar bounds as Theorem~\ref{thm:MI_pos_acc}. We give a practical counter-example that shows for a specific MI attack on an $(\varepsilon,\delta)$-DP logistic regression there is no bound on its positive accuracy, unlike what follows from our bound for $\varepsilon$-DP; note, however, that we do have bounds on MI accuracy (as opposed to positive accuracy) when using $(\varepsilon,\delta)$-DP as seen by \cite{yeom2018privacy,erlingsson2019we}.

Following this, we provide a bound on MI positive accuracy when using $(\varepsilon,\delta)$-DP, but the bound introduces a necessary $\mathbb{P}(S| \mathbf{x}^*)$ dependence. We say necessary as the previous example showed MI positive accuracy becomes unbounded as $\mathbb{P}(S| \mathbf{x}^*)$ becomes small.

\subsection{Positive Accuracy is not always bounded}
\label{sssec:pos_acc_not_bnd}
Consider a set of two (scalar) points $\{x_1, x_2\}$ which are drawn into the training set $D$ with $\mathbb{P}_{x_1}(1) = 1$ and $\mathbb{P}_{x_2}(1) = 0.5$; that is $x_1$ is always in the training set, and $x_2$ has a $50\%$ chance of being in the training set.
Let model $M$ be a single dimensional logistic regression without bias defined as $M(x) = wx$ initialized such that for cross-entropy loss $L$, $\nabla L |_{\mathbf{x}_1} \approx 0$ (\ie set $x_1 = \{(10^6,1)\}$ and the weights $w = 10^6$ so that $M(x) = 10^6 x$ and thus the softmax output of the model is approximately $1$ on $x_1$ and thus gradient is approximately $0$). Conversely set $x_2$ such that the gradient on it is less than $-1$ (\ie for the above setting set $x_2 = \{(10^6,0)\}$).

Now, train the model to $(1,10^{-5})$-DP following \citet{abadi2016deep} with $\eta = 1$, sampling rate of $100\%$, a maximum gradient norm of $1$, for one step. Note these are just parameters which we are allowed to change under the DP analysis, and we found the noise we would need for $(1,10^{-5})$-DP is $4.0412$.
Then consider running a MI attack on the above setup where if for some threshold $\alpha$ the final weights $\mathbf{W}$ are s.t if $\mathbf{W} \leq \alpha$ one classifies those weights as having come from the dataset with $x_2$, otherwise not. Do note that here we use $\mathbf{W}$ for the final weights as opposed to $w$ to emphasize that we are now talking about a random variable. The intuition for this attack is that if the dataset only contain $x_1$ then the weights do not change, but if the dataset contains $x_2$ we know the resulting gradient is negative (by construction) and thus decreases the weights (before noise).

By the earlier setting on how training datasets are constructed note that $D = \{x_1\}$ or $D = \{x_1,x_2\}$, and we will denote these $D_1$ and $D_2$ respectively. Note that if $M$ trained on $D_1$ following the suggested data points and initial weights, we have the distribution of final weight $\mathbf{W}_{D_1} = N(10^6,\sigma) = N(10^6,4.0412)$ where $\sigma$ denotes the noise needed for $(1,10^{-5})$-DP as stated earlier. Similarly $\mathbf{W}_{D_2} = N(10^6-1,4.0412)$, since the maximum gradient norm is set to 1.

\begin{figure}
     \centering
     \begin{subfigure}[b]{0.48\textwidth}
         \centering
         \includegraphics[width=70mm]{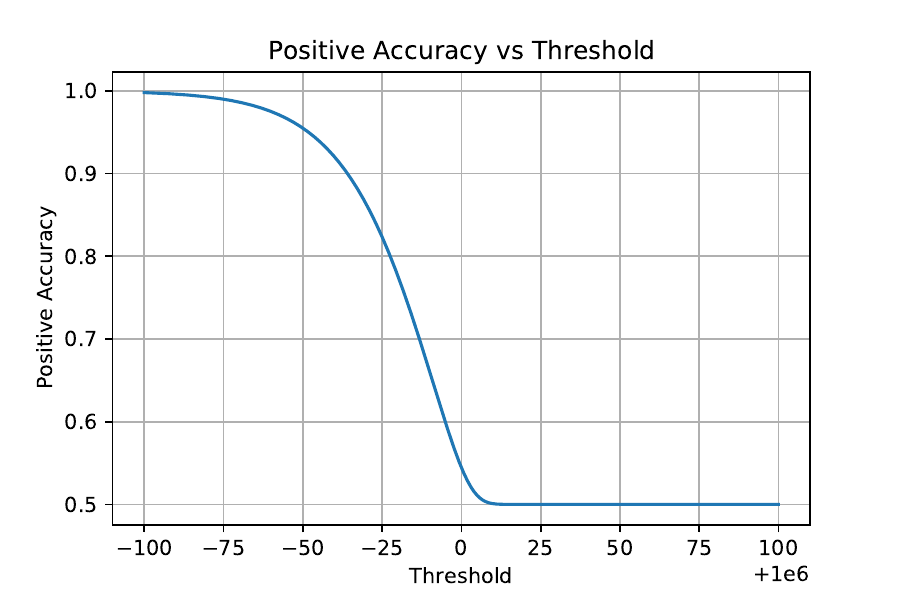}
         \caption{Positive accuracy as a function of the threshold}
         \label{fig:threshold_pos_acc}
     \end{subfigure}
     \hfill
     \begin{subfigure}[b]{0.48\textwidth}
         \centering
         \includegraphics[width=70mm]{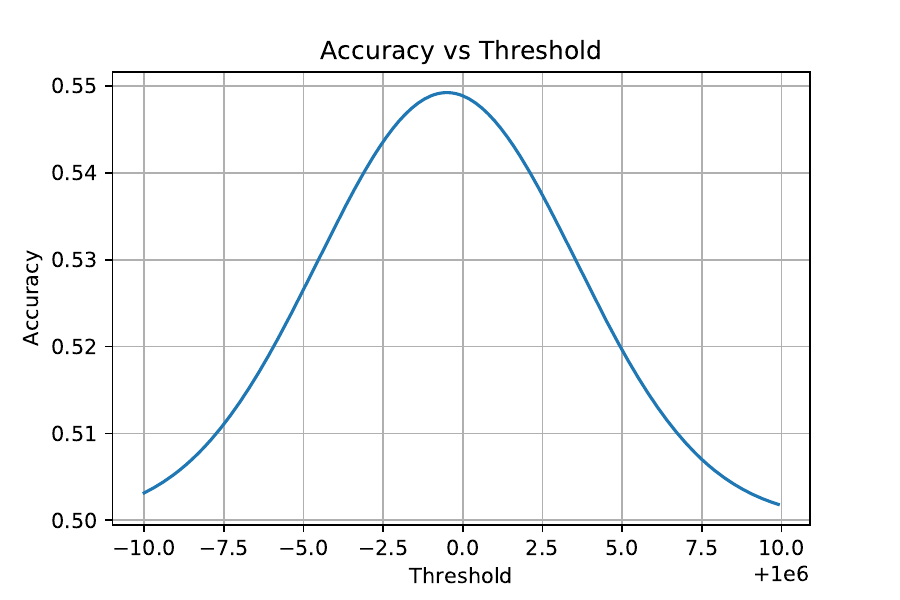}
         \caption{Accuracy as a function of the threshold}
         \label{fig:threshold_acc}
     \end{subfigure}
        \caption{Impact of threshold on positive accuracy and accuracy.}
        \label{fig:func_thresh}
\end{figure}

For the above MI attack we can then bound the positive accuracy  as a function of $\alpha$ by: $$\mathbb{P}(D_2|\mathbf{W}\leq \alpha) \\
=  \frac{\mathbb{P}(\mathbf{W}_{D_2} \leq \alpha)*\mathbb{P}(D_2)}{\mathbb{P}(\mathbf{W}_{D_2} \leq \alpha)*\mathbb{P}(D_2) + \mathbb{P}(\mathbf{W}_{D_1} \leq \alpha)*\mathbb{P}(D_1)}$$
$$= \frac{\phi(\mathbf{W}_{D_2},\alpha)*0.5}{\phi(\mathbf{W}_{D_1},\alpha)*0.5 + \phi(\mathbf{W}_{D_2},\alpha)*0.5}.$$ Note $\phi(\mathbf{W},\alpha)$ is the (Gaussian) cumulative function of random variable $\mathbf{W}$ upto $\alpha$. We plot this in~\Cref{fig:threshold_pos_acc}, and unlike Theorem~\ref{thm:MI_pos_acc}, note how it is not bounded by anything less than $1$ and goes to $1$ as the threshold $\alpha$ decreases (\ie $\forall m \in [0,1)~\exists \alpha$ s.t $S = (-\infty,\alpha]$ yields positive accuracy greater than $m$).

\subsection{An $(\varepsilon, \delta)$-DP Bound}

We have seen so far that MI positive accuracy is not always bounded for $(\varepsilon,\delta)$-DP. Nevertheless, Theorem~\ref{thm:MI_pos_acc} can be generalized to $(\varepsilon,\delta)$-DP if we introduce a $\mathbb{P}(S|\mathbf{x}^*)$ dependence.

\begin{theorem}
\label{thm: eps_delta_bound}
\new{For any MI attack on datapoint $\mathbf{x}^*$ (\ie with any associated set $S$)}, and the training process $H$ is DP with $(\varepsilon, \delta)$, its \new{positive} accuracy is upper-bounded by $\mathbb{P}(\mathbf{x}^* | S) \leq (1 + \frac{e^{-\varepsilon}\mathbb{P}_{\mathbf{x}^*}(0)}{\mathbb{P}_{\mathbf{x}^*}(1)} - \frac{\delta e^{-\varepsilon}\mathbb{P}_{\mathbf{x}^*}(0)}{\mathbb{P}(S | \mathbf{x}^*)})^{-1}$ \new{so long as the right-hand side is positive}.
\end{theorem}

The proof follows the same strategy as Theorem~\ref{thm:MI_pos_acc}, but also multiplies by $1$ in a particular way to deal with the $\delta$ terms. Do note however that some form of a $\mathbb{P}(S| \mathbf{x}^*)$ dependence in the bound is necessary since the MI positive accuracy, as shown in a previous subsection, becomes unbounded as $\mathbb{P}(S| \mathbf{x}^*)$ gets arbitrarily small. That is, for there to exist a bound on MI positive accuracy, we need a lower bound on $\mathbb{P}(S| \mathbf{x}^*)$; the bound in Theorem~\ref{thm: eps_delta_bound} shows this is in fact sufficient.

\begin{proof}

Following the proof of Theorem 1, we have the positive accuracy of any adversary $f$ \new{(\ie associated to any set $S$)} is:

\begin{equation}
    \mathbb{P}(\mathbf{x}^*|S) = \frac{\sum_{D \in \mathfrak{D}_{\mathbf{x}^*}}\mathbb{P}(H(D) \in S)\mathbb{P}(D)}{\sum_{D \in \mathfrak{D}_{\mathbf{x}^*}}\mathbb{P}(H(D) \in S)\mathbb{P}(D) + \sum_{D' \in \mathfrak{D}_{\mathbf{x}^*}'}\mathbb{P}(H(D') \in S)\mathbb{P}(D')}
\end{equation}

Now using that $\mathbb{P}(H(D') \in S) \geq e^{-\varepsilon}\mathbb{P}(H(D) \in S) - e^{-\varepsilon} \delta$ and $\mathbb{P}(D') = \mathbb{P}(D) \frac{\mathbb{P}_{\mathbf{x}^*}(0)}{\mathbb{P}_{\mathbf{x}^*}(1)}$ we have the denominator is $\geq \sum_{D \in \mathfrak{D}_{\mathbf{x}^*}} (1 + \frac{e^{-\varepsilon}\mathbb{P}_{\mathbf{x}^*}(0)}{\mathbb{P}_{\mathbf{x}^*}(1)})\mathbb{P}(D) - \\ \delta e^{-\varepsilon} \sum_{D' \in \mathfrak{D}_{\mathbf{x}^*}'}\mathbb{P}(D')$.

Note that $$\delta e^{-\varepsilon} \sum_{D' \in \mathfrak{D}_{\mathbf{x}^*}'}\mathbb{P}(D') = \delta e^{-\varepsilon} \sum_{D' \in \mathfrak{D}_{\mathbf{x}^*}'}\mathbb{P}(D') \\ \frac{\sum_{D \in \mathfrak{D}_{\mathbf{x}^*}}\mathbb{P}(H(D) \in S)\mathbb{P}(D)}{\sum_{D \in \mathfrak{D}_{\mathbf{x}^*}}\mathbb{P}(H(D) \in S)\mathbb{P}(D)}.$$ So now cancelling the $\sum_{D \in \mathfrak{D}_{\mathbf{x}^*}} \mathbb{P}(H(D)\in  S)\mathbb{P}(D)$ in the numerator and denominator, and noting $ \sum_{D \in \mathfrak{D}_{\mathbf{x}^*}}  \mathbb{P}(H(D) \in S)\mathbb{P}(D) = \mathbb{P}(S| \mathbf{x}^*)$ and $\sum_{D' \in \mathfrak{D}_{\mathbf{x}^*}'}\mathbb{P}(D') = \mathbb{P}_{\mathbf{x}^*}(0)$, we have $\mathbb{P}(\mathbf{x}^*|S) \leq (1 + \frac{e^{-\varepsilon}\mathbb{P}_{\mathbf{x}^*}(0)}{\mathbb{P}_{\mathbf{x}^*}(1)} - \frac{\delta e^{-\varepsilon}\mathbb{P}_{\mathbf{x}^*}(0)}{\mathbb{P}(S | \mathbf{x}^*)})^{-1}$. \new{Note this is only true if the right-hand side of the bound is positive, as the fact $a/b \leq a/c$ if $c \leq b$ (which we used) is only true if $c$ is positive when $a$ and $b$ is positive}

\end{proof}

\paragraph{Remarks on $(\varepsilon,\delta)$-DP}

As shown in Theorem~\ref{thm: eps_delta_bound}, we can obtain bounds on MI positive accuracy, but we must accept a probability of failure. This is not the case in bounds on MI accuracy with $(\varepsilon,\delta)$-DP, illustrating a difference between the formal study of the two. We will proceed to focus on $\varepsilon$-DP in the other sections of the paper, however everything follows analogously for $(\varepsilon,\delta)$-DP if one lower-bounds $\mathbb{P}(S|\mathbf{x}^*)$, \ie accepts $X$ probability of failure, and so only studies $\mathbb{P}(S| \mathbf{x}^*) > X$.

\subsection{Comparison with $(1-\delta)$ Approach to $(\varepsilon,\delta)$ Positive Accuracy Bounds}
\label{appendix: alt_failure}

As a last point of discussion on $(\varepsilon,\delta)$-DP, we provide a comparison to \citet{mahloujifar2022optimal} and outline how our analysis yields a more relevant (and stronger)
approach to bounding failure rates when using $(\varepsilon,\delta)$-DP. Following Eq. 15 in \citet{mahloujifar2022optimal} which states the $\varepsilon$-DP conditions holds for any fixed $(D,D')$ pair with probability $(1-\delta)$, the bounds in Theorem \ref{thm:MI_pos_acc} trivially apply to $(\varepsilon,\delta)$-DP with probability $(1 -(\delta))^N$ where $N$ is the number of $(D,D’)$ pairs appearing from the random sampling. Note this $N$ could be quite large, \ie if $n$ is the number of datapoints in the larger set, $N= \sum_{i =1}^{n-1} {n-1 \choose j}$ if every point is being randomly sampled. The scenario studied by \citet{mahloujifar2022optimal} bypasses the $N$ dependence by only considering one $(D,D’)$ pair, \ie~all the other points in the dataset are fixed and only the datapoint in question is being randomly sampled. This assumption made by \citet{mahloujifar2022optimal} is not applicable in practice as we do not know which datapoints the adversary is going to test: hence, we would have to resort to randomly sampling all of them. Instead, our failure analysis in Section~\ref{sec:eps_delta} (particularly Theorem~\ref{thm: eps_delta_bound}) conveniently bypasses this problem by telling us low probability states are what lead to failures (removing the $N$ dependence in the failure rate).

\section{The Effect of $\mathbb{P}_{\mathbf{x}^*}(1)$}
\label{sec: sampling_effect}

We now focus on the effect of $\mathbb{P}_{\mathbf{x}^*}(1)$. Section~\ref{ssec:priv_amp} explains how the privacy amplification, \ie~lowering of our MI positive accuracy bound, we observe from decreasing $\mathbb{P}_{\mathbf{x^*}}(1)$ is fundamentally different than the privacy amplification on MI from batch sampling. Section~\ref{ssec:usefulness_defender} outlines the practical consequences of this for a defender.

\subsection{A New Privacy Amplification for MI}
\label{ssec:priv_amp}

Our bound given by Theorem \ref{thm:MI_pos_acc} can be reduced by decreasing $\mathbb{P}_{\mathbf{x}^*}(1)$ or $\varepsilon$. Furthermore, note batch sampling, the probability for a data point to be used in the batch for a given training step, reduces MI positive accuracy as it reduces $\varepsilon$. So dataset sub-sampling ($\mathbb{P}_{\mathbf{x}^*}(1)$) and batch sampling both decrease our bound, and we term their effect "privacy amplification" (for MI) as they decrease privacy infringement (analogous to how "privacy amplification" for DP refers to methods that reduce privacy loss). However, is the effect of dataset sub-sampling and batch sampling different?

\begin{figure}[h]
    \centering
    \includegraphics[width=70mm]{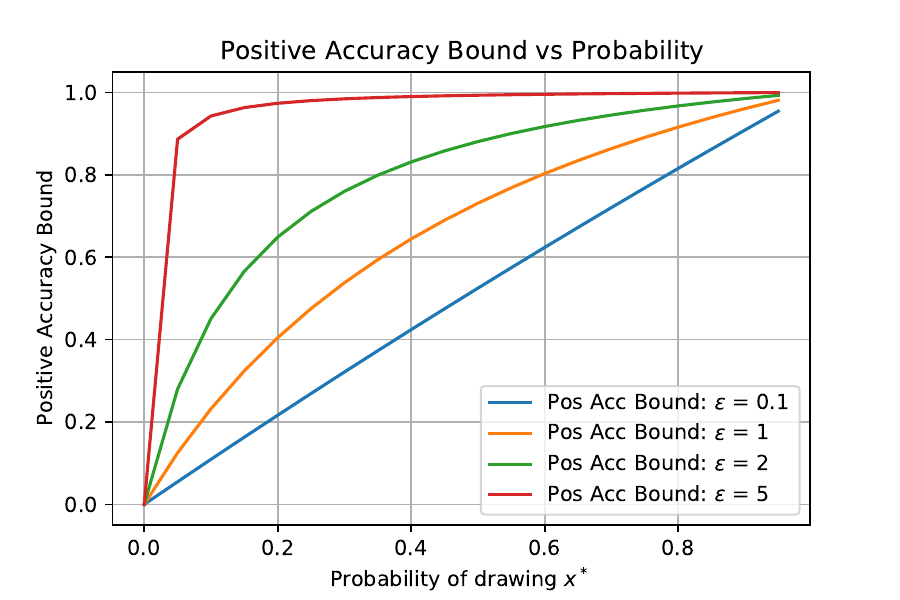}
    \caption{Our upper bound on MI positive accuracy as a function of $\mathbb{P}_{\mathbf{x}^*}(1)$}
    \label{fig:MI_bound_prob}
\end{figure}

Before proceeding, it is useful to get a sense of the impact $\mathbb{P}_{\mathbf{x}^*}(1)$ has on our bound. We plot $\mathbb{P}_{\mathbf{x}^*}(1)$ against our positive MI accuracy bound given by Theorem \ref{thm:MI_pos_acc} in~\Cref{fig:MI_bound_prob} for different $\varepsilon$. Notably, for a specific case when $\mathbb{P}_{\mathbf{x}^*}(1)$ is small, we get that the positive accuracy is bounded by $6.9\%$ for $\varepsilon=2$ and $\mathbb{P}_{\mathbf{x}^*}(1) = 0.01$ (\ie by sampling with low enough probability, we certify that they will be correct at most $6.9\%$ of the time).

\begin{figure}[h]
    \centering
    \includegraphics[width=70mm]{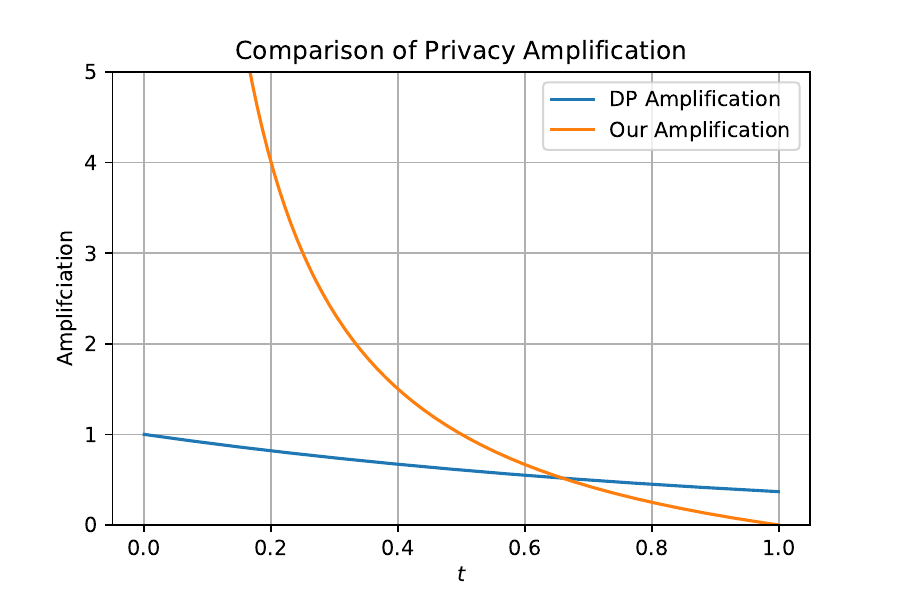}
    \caption{Comparing the DP amplification observed by decreasing batch probability (given by $e^{-t}$) to the amplification we observe from decreasing $\mathbb{P}_{\mathbf{x}}(1)$ (given by $(1-t)/t$).}
    \label{fig:Priv_Amp_Comp}
\end{figure}

We now turn to comparing the effect of batch sampling to the effect of $\mathbb{P}_{\mathbf{x}^*}(1)$ (note $\mathbb{P}_{\mathbf{x}^*}(0)/\mathbb{P}_{\mathbf{x}^*}(1) = (1- \mathbb{P}_{\mathbf{x}^*}(1))/\mathbb{P}_{\mathbf{x}^*}(1)$). First it is worth noting that the two amplification methods are mostly independent, \ie decreasing $\mathbb{P}_{\mathbf{x}^*}(0)/\mathbb{P}_{\mathbf{x}^*}(1)$ mostly places no restriction on the sampling rate for batch sizes (with some exception).\footnote{We say "mostly" as this is true upto a point. In particular the expectation of the training dataset size decreases with smaller dataset sampling probabilities, and thus the lowest possible batch sampling rate $1/n$ increases in expectation.} Nevertheless we can ignore this restriction for the time being as we are interested in their independent mathematical behaviours. Including $q$ for DP batch privacy amplification, we can compare its impact to $\mathbb{P}_{\mathbf{x}^*}(1)$ by looking at the term $e^{-q\varepsilon_0}\mathbb{P}_{\mathbf{x}}(0)/\mathbb{P}_{\mathbf{x}}(1)$ in the bound given by Theorem \ref{thm:MI_pos_acc}; the goal is to maximize this to make the upper bound as small as possible. In particular, we see that decreasing $q$ increases this term by $O(e^{-t})$ where as decreasing $\mathbb{P}_{\mathbf{x}^*}(1)$ increases this term by $O((1-t)/t)$, which is slower than $O(e^{-t})$ up to a point, then faster (do note that we are looking at the order as the variable $t$ decreases). Figure~\ref{fig:Priv_Amp_Comp} plots this relation, however note that the specific values are subject to change with differing constant. Nevertheless what does not change with the constants are the asymptotic behaviours, and in particular we see $\lim_{t \rightarrow 0} O((1-t)/t) = \infty$ where as $\lim_{t \rightarrow 0} O(e^{-t}) =~ \text{constant}$.

Thus, we can conclude the effects of data sampling and batch sampling are different to our bound. Therefore, data sampling presents a complementary privacy amplification scheme for MI positive accuracy. As a last remark, note that the same comparison holds more generally when comparing the effect of $\varepsilon$ and $\mathbb{P}_{\mathbf{x}^*}(1)$. %

\subsection{Usefulness for a Defender}
\label{ssec:usefulness_defender}

We now explain one course of action a defender can take in light of this new privacy amplification for MI. In particular note that an upper bound on $\mathbb{P}_{\mathbf{x}^*}(1)$ translates to an upper bound on the relation found in Theorem \ref{thm:MI_pos_acc} (as the bound is monotonically increasing with $\mathbb{P}_{\mathbf{x}^*}(1)$); hence one can, in practice, focus on giving smaller upper-bounds on $\mathbb{P}_{\mathbf{x}^*}(1)$ to decrease MI positive accuracy.

A possible approach to this can be described as follows: assume a user (the defender) is given some sample $D \subset \{\mathbf{x}_1,\cdots,\mathbf{x}_N\}$ drawn with some unknown distribution. In particular, the user does not know the individual probabilities for the points being in $D$ and whether these points were drawn independently. However, assume that the user obtains from $D$ the training dataset $D_{train}$ by sampling any point from $D$ independently with probability $T$. Our bound then bounds the conditional probability $\mathbb{P}(\mathbf{x}^*|S, \mathfrak{D})$ with $\mathbb{P}_{\mathbf{x}^*}(1) = T$. Note that $\mathbb{P}(\mathbf{x}^*|S) \leq \mathbb{P}(\mathbf{x}^*|S,\mathfrak{D})$ as $\mathbf{x}^* \in \mathfrak{D}$, so our upper-bound also bounds $\mathbb{P}(\mathbf{x}^*|S)$. That is, one can use our bound to give upper-bounds for points sampled from unknown (not necessarily i.i.d) distributions if independent sub-sampling (\ie, as needed for Lemma 1) is applied \textit{after} drawing from this unknown distribution.

This does come with some drawbacks. In general one wants to train with more data, but by further sampling with probability $T$ we reduce our expected training dataset size. As a consequence, a user will have to make the decision between how low they can make $T$ (in conjunction with the $\varepsilon$ parameter they choose) compared to how small a dataset they are willing to train on. We leave this type of decision making for future work.

\subsection{Empirical Attack Case Study}

To explore the practical effect of $\mathbb{P}_{\mathbf{x}^*}(1)$ on MI positive accuracy when training with different $(\varepsilon,\delta)$-DP guarantees, we implement the current state-of-the-art membership inference attack: LiRA~\cite{carlini2022membership}. In the original paper, the attack was not evaluated against DP trained models, so we reimplement the attack ourselves on MNIST and CIFAR10. In particular, we follow Algorithm 1 in~\citet{carlini2022membership} but now the training function is DP-SGD, varying $\varepsilon$ with fixed $\delta = 10^{-5}$, and using the hyperpaprameters of~\cite{tramer2020differentially} to train the models. We train 20 shadow models for each $\varepsilon$ setting (\ie~reuse the same shadow models for all the points we test), sampling their training sets uniformly with $50\%$ probability from the original training set. Our target models use the same training function as the shadow models, but we now vary the $\mathbb{P}_{x^*}(1)$ to capture the effectiveness of the attack on target models with different sampling rates.

We evaluate the attack by computing the scores (outputted by Algorithm 1~\cite{carlini2022membership}) for the first $1000$ training points of MNIST and CIFAR10 respectively. In each setting we consider different thresholds for acceptance according to a quantile based approach: we consider accepting the top $i*2.5\%$ quantile of scores for every $i \in [40]$, and report the max precision (positive accuracy) of all these different thresholds (\ie~select the highest precision threshold).

\begin{figure}
     \centering
     \begin{subfigure}[b]{0.4\textwidth}
         \centering
         \includegraphics[width=\textwidth]{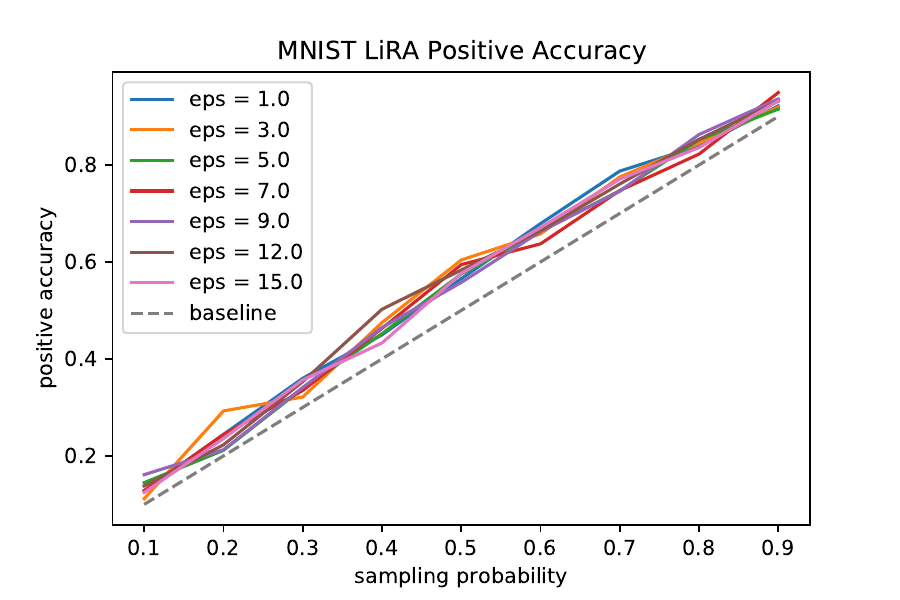}
         \caption{MNIST LiRA positive accuracy for different $\varepsilon$ as a function of $\mathbb{P}_{\mathbf{x}^*}(1)$}
     \end{subfigure}
     \hfill
     \begin{subfigure}[b]{0.4\textwidth}
         \centering
         \includegraphics[width=\textwidth]{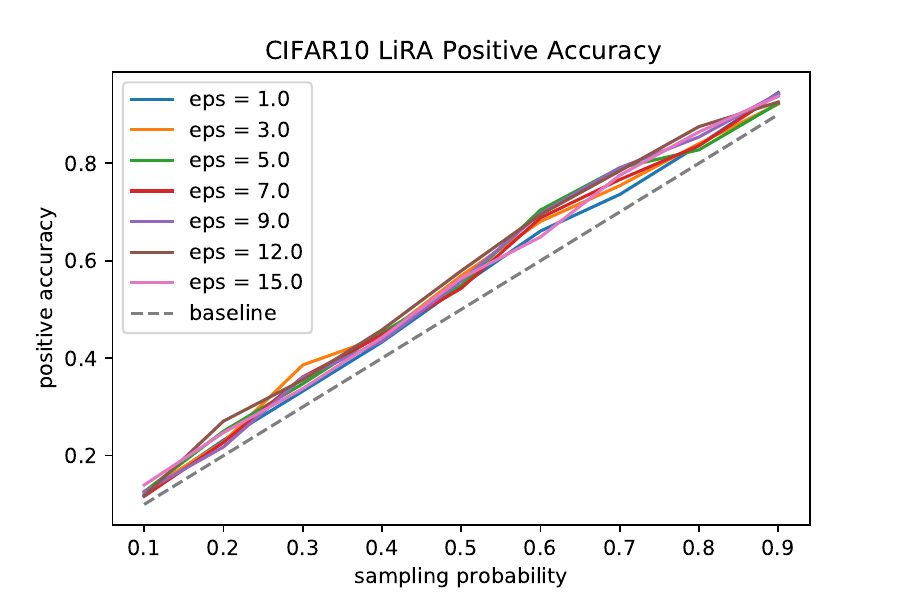}
         \caption{CIFAR10 LiRA positive accuracy for different $\varepsilon$ as a function of $\mathbb{P}_{\mathbf{x}^*}(1)$}
     \end{subfigure}
        \caption{Plots of the positive accuracy of the LiRA attack on MNIST and CIFAR10}
        \label{fig:LiRA_pos_acc}
\end{figure}

Figure~\ref{fig:LiRA_pos_acc} plots the average positive accuracy (from $5$ repeated trials) of the LiRA attack for different $\varepsilon$ as a function of the $\mathbb{P}_{\mathbf{x}^*}(1)$ used for the target model.  We further plot the positive accuracy of the baseline attack of always predicting a datapoint is in the training dataset. As we see the LiRA attack consistently outperforms the baseline, but importantly, \textbf{we see that the sampling is the dominant factor dictating how much we outperform the baseline.}

\begin{figure}
     \centering
     \begin{subfigure}[b]{0.4\textwidth}
         \centering
         \includegraphics[width=\textwidth]{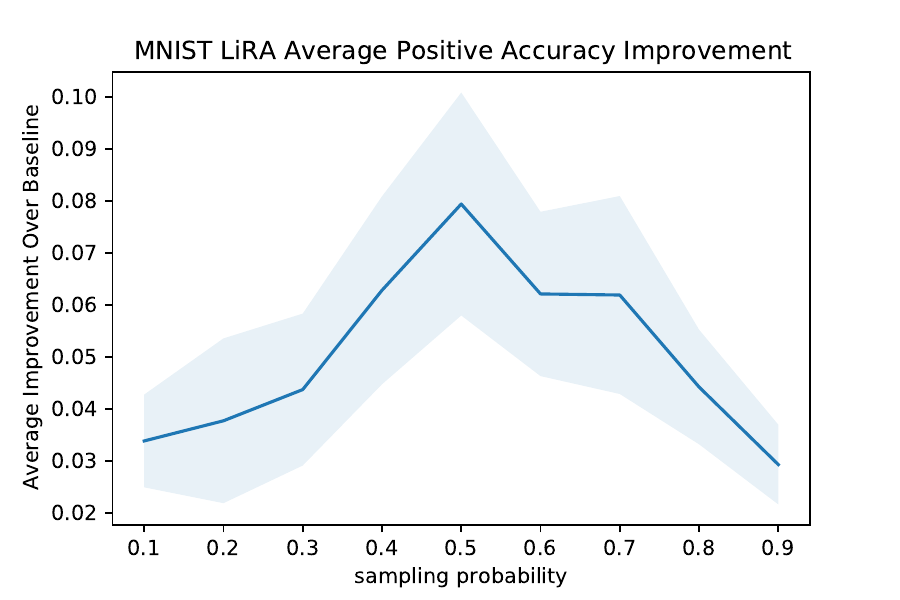}
         \caption{MNIST LiRA average positive accuracy (across $\varepsilon$) improvement against the baseline attack}
     \end{subfigure}
     \hfill
     \begin{subfigure}[b]{0.4\textwidth}
         \centering
         \includegraphics[width=\textwidth]{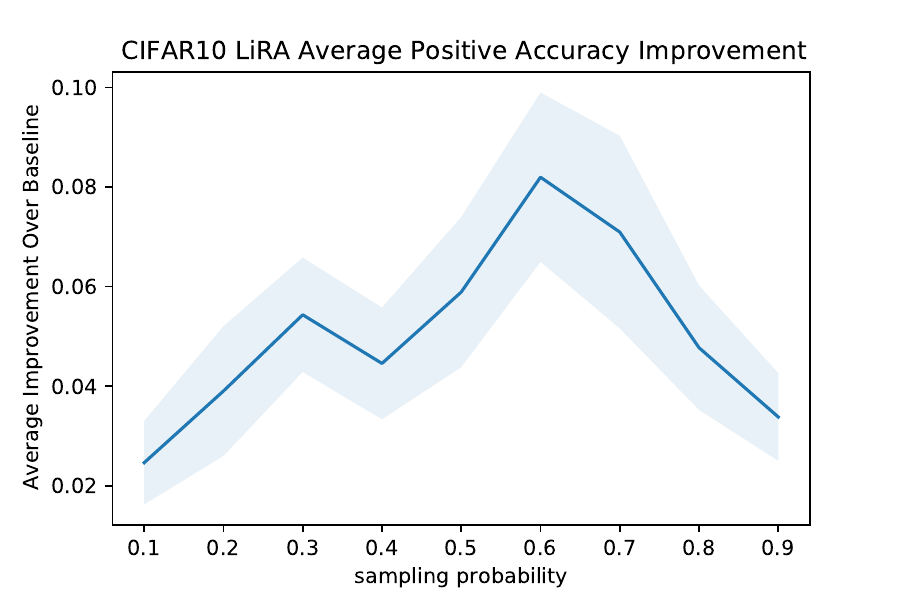}
         \caption{CIFAR10 LiRA average positive accuracy (across $\varepsilon$) improvement against the baseline attack}
     \end{subfigure}
        \caption{Plots of the positive accuracy improvement of the LiRA attack over the baseline attack on MNIST and CIFAR10. We include the $95\%$ confidence interval (over the $epsilon$ and trials).}
        \label{fig:LiRA_improvement}
\end{figure}

To illustrate this outweighed effect of sampling over $\varepsilon$, Figure~\ref{fig:LiRA_improvement} plots the average improvement (over $\varepsilon$ and trials) against the baseline attack. What we see is that both low and high sampling rates see less of an improvment over the baseline than medium sampling rates. In particular, this shows \emph{low sampling disproportionately mitigates membership inference}. This behaviour is largely consistent with what we expect from our bounds (see Figure~\ref{fig:MI_bound_prob}).

\subsection{Effect on Accuracy}

\begin{figure}
     \centering
     \begin{subfigure}[b]{0.3\textwidth}
         \centering
         \includegraphics[width=\textwidth]{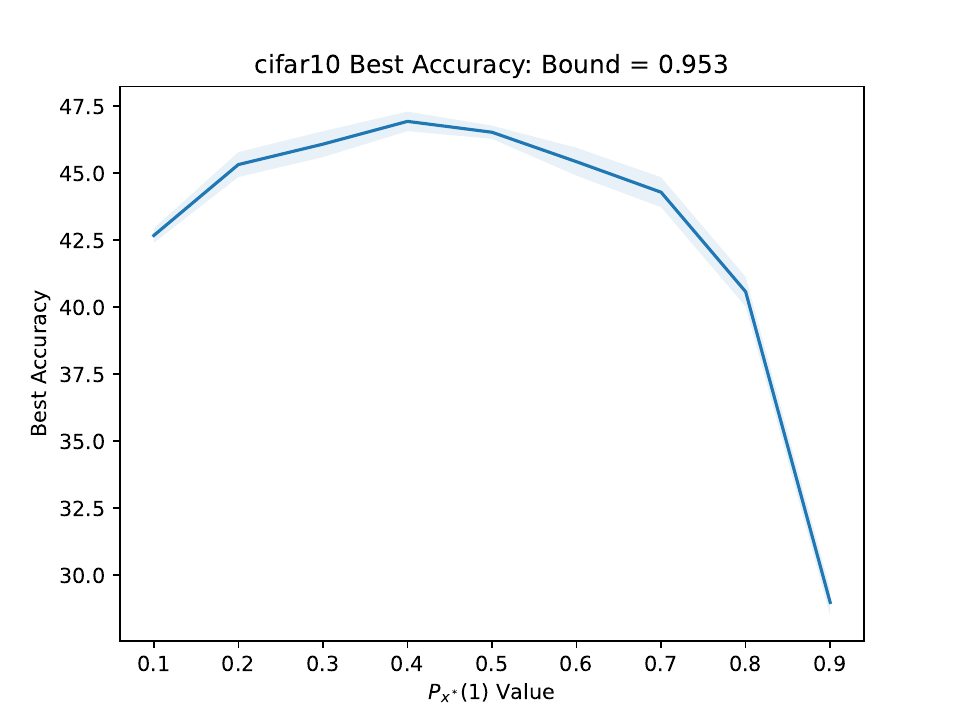}
         \caption{$\varepsilon=3$ when $\mathbb{P}_{\mathbf{x}^*}(1) = 0.5$}
     \end{subfigure}
     \hfill
     \begin{subfigure}[b]{0.3\textwidth}
         \centering
         \includegraphics[width=\textwidth]{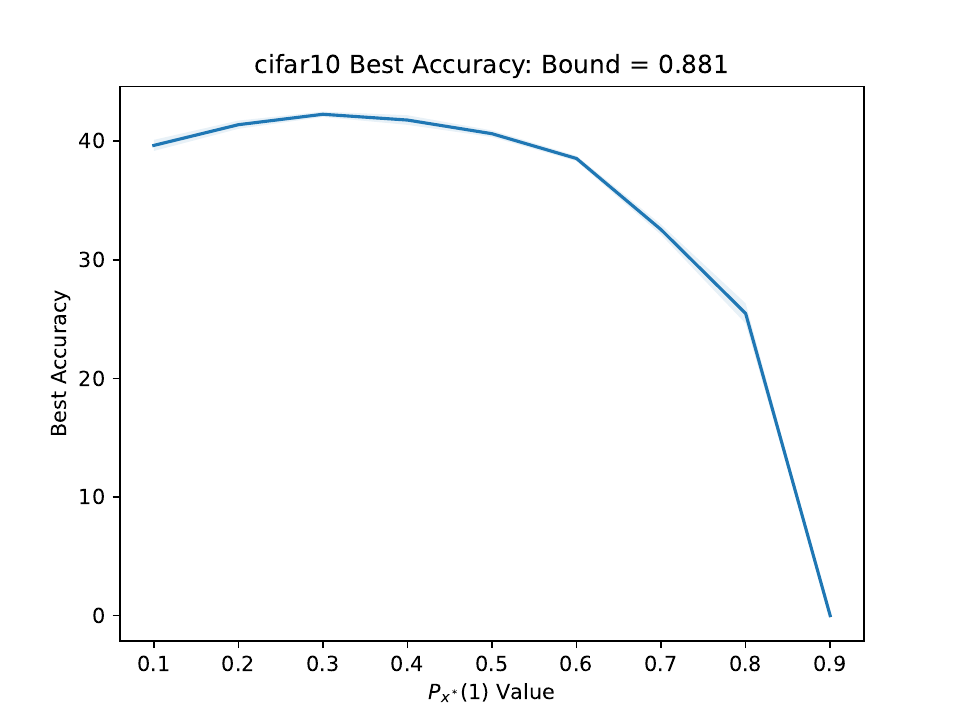}
         \caption{$\varepsilon=2$ when $\mathbb{P}_{\mathbf{x}^*}(1) = 0.5$}
     \end{subfigure}
     \hfill
     \begin{subfigure}[b]{0.3\textwidth}
         \centering
         \includegraphics[width=\textwidth]{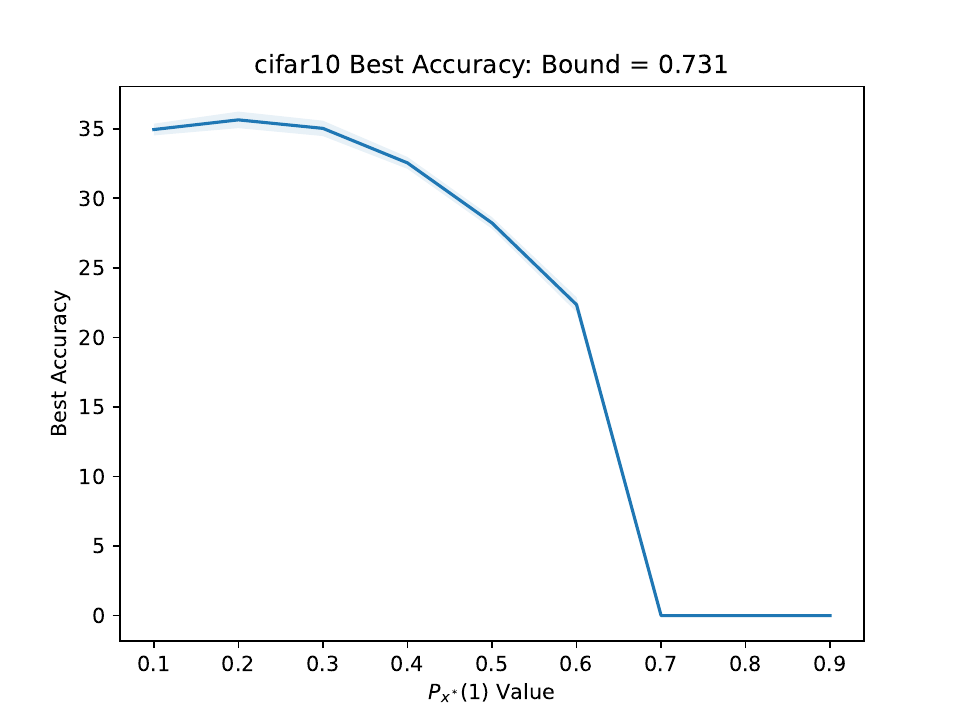}
         \caption{$\varepsilon=1$ when $\mathbb{P}_{\mathbf{x}^*}(1) = 0.5$}
         \label{fig:five over x}
     \end{subfigure}
        \caption{Plots of the best accuracy vs. $\mathbb{P}_{\mathbf{x}^*}(1)$ at differing fixed bounds for CIFAR10. We include the $95\%$ confidence interval (over the trials).}
        \label{fig:best_acc_cifar10}
\end{figure}

\begin{figure}
     \centering
     \begin{subfigure}[b]{0.3\textwidth}
         \centering
         \includegraphics[width=\textwidth]{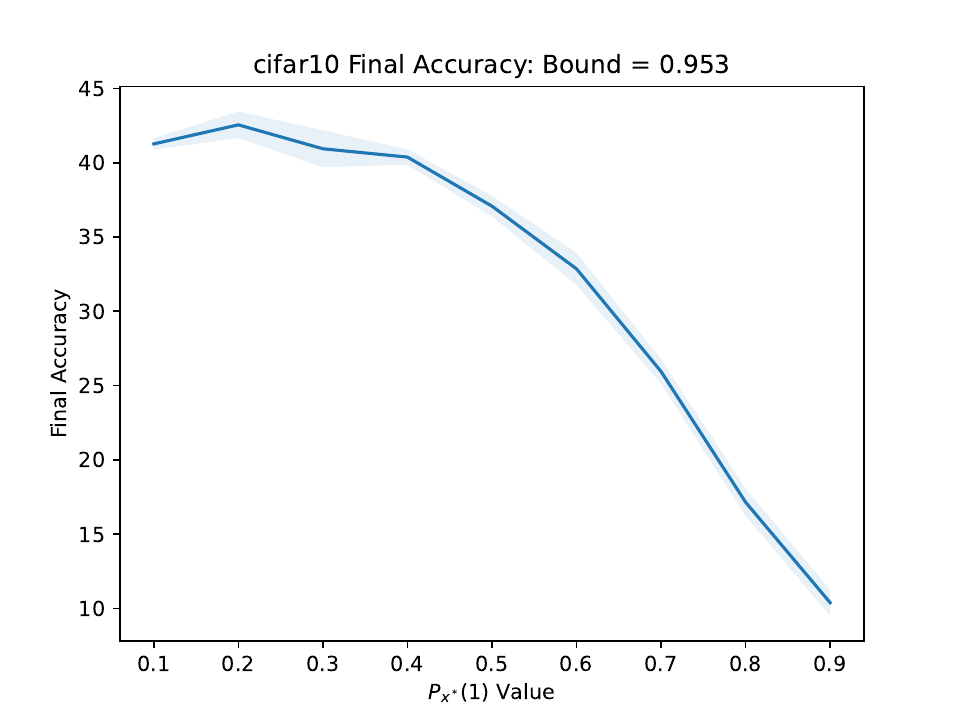}
         \caption{$\varepsilon=3$ when $\mathbb{P}_{\mathbf{x}^*}(1) = 0.5$}
     \end{subfigure}
     \hfill
     \begin{subfigure}[b]{0.3\textwidth}
         \centering
         \includegraphics[width=\textwidth]{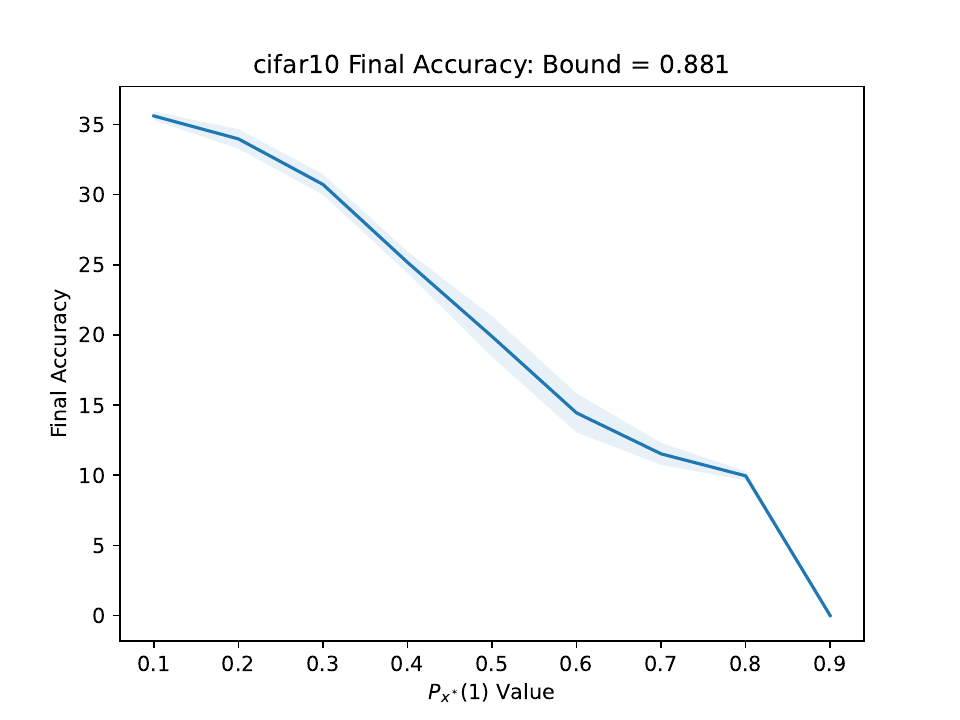}
         \caption{$\varepsilon=2$ when $\mathbb{P}_{\mathbf{x}^*}(1) = 0.5$}
     \end{subfigure}
     \hfill
     \begin{subfigure}[b]{0.3\textwidth}
         \centering
         \includegraphics[width=\textwidth]{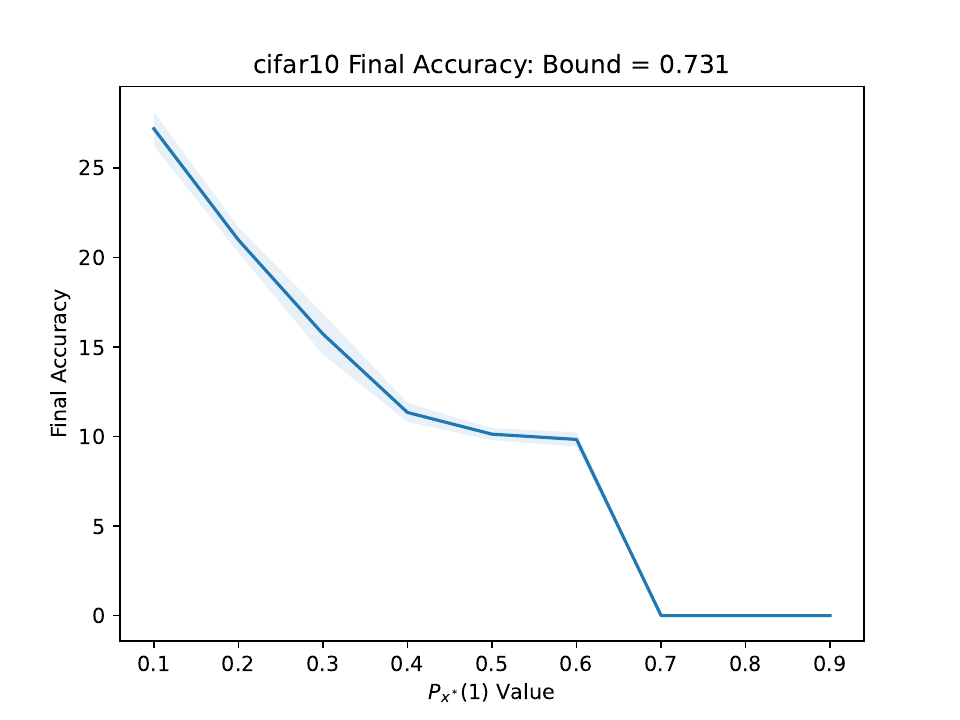}
         \caption{$\varepsilon=1$ when $\mathbb{P}_{\mathbf{x}^*}(1) = 0.5$}
         \label{fig:five over x}
     \end{subfigure}
        \caption{Plots of the final accuracy vs. $\mathbb{P}_{\mathbf{x}^*}(1)$ at differing fixed bounds for CIFAR10. We include the $95\%$ confidence interval (over the trials).}
        \label{fig:final_acc_cifar10}
\end{figure}

\begin{figure}
     \centering
     \begin{subfigure}[b]{0.3\textwidth}
         \centering
         \includegraphics[width=\textwidth]{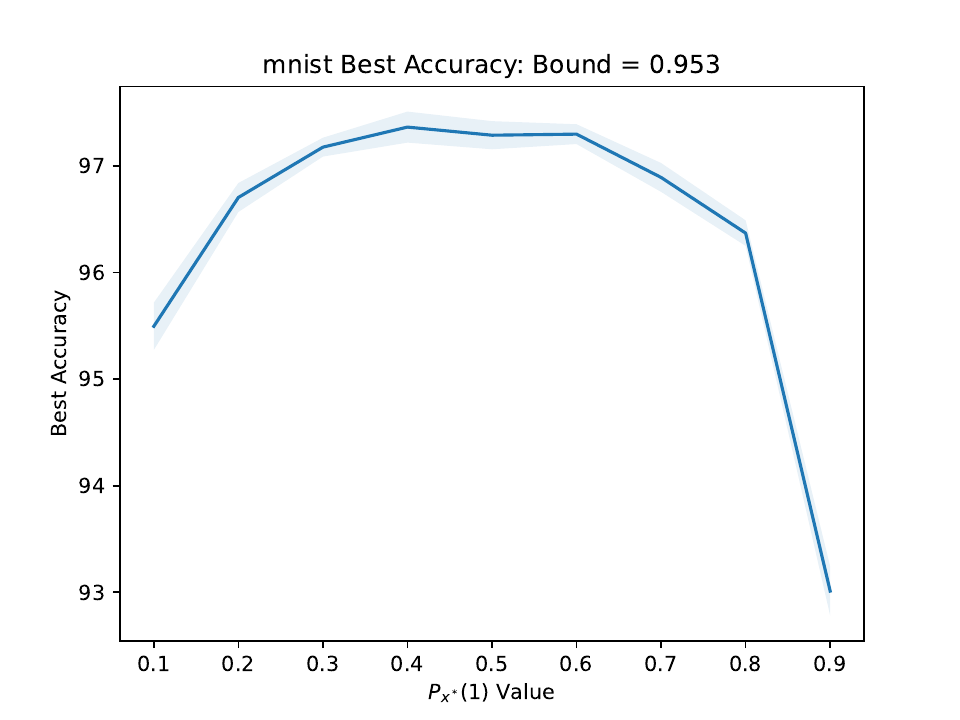}
         \caption{$\varepsilon=3$ when $\mathbb{P}_{\mathbf{x}^*}(1) = 0.5$}
     \end{subfigure}
     \hfill
     \begin{subfigure}[b]{0.3\textwidth}
         \centering
         \includegraphics[width=\textwidth]{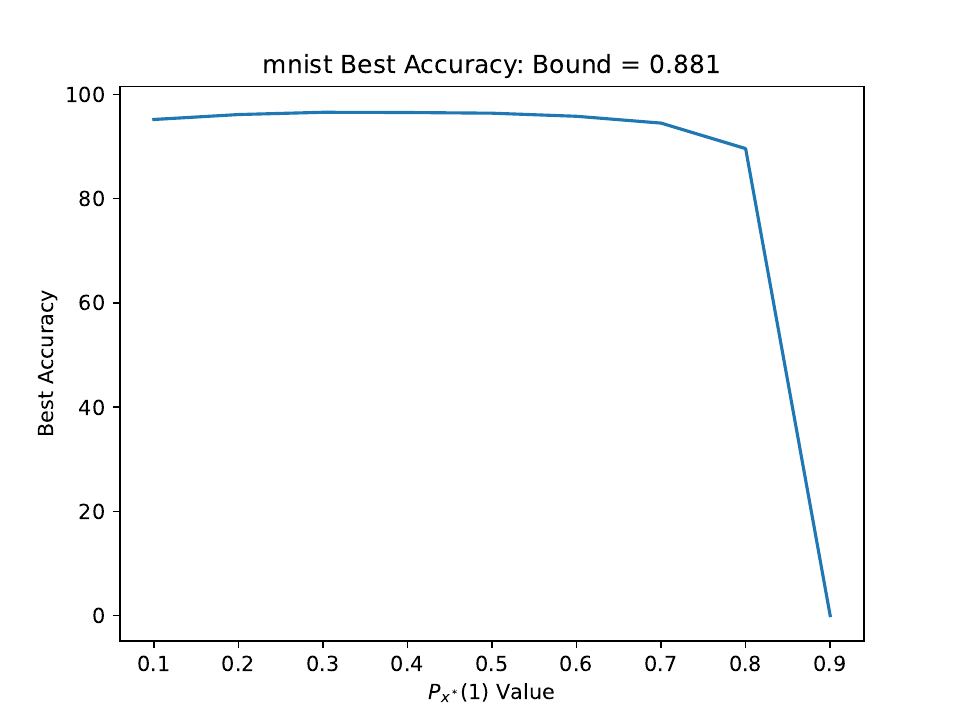}
         \caption{$\varepsilon=2$ when $\mathbb{P}_{\mathbf{x}^*}(1) = 0.5$}
     \end{subfigure}
     \hfill
     \begin{subfigure}[b]{0.3\textwidth}
         \centering
         \includegraphics[width=\textwidth]{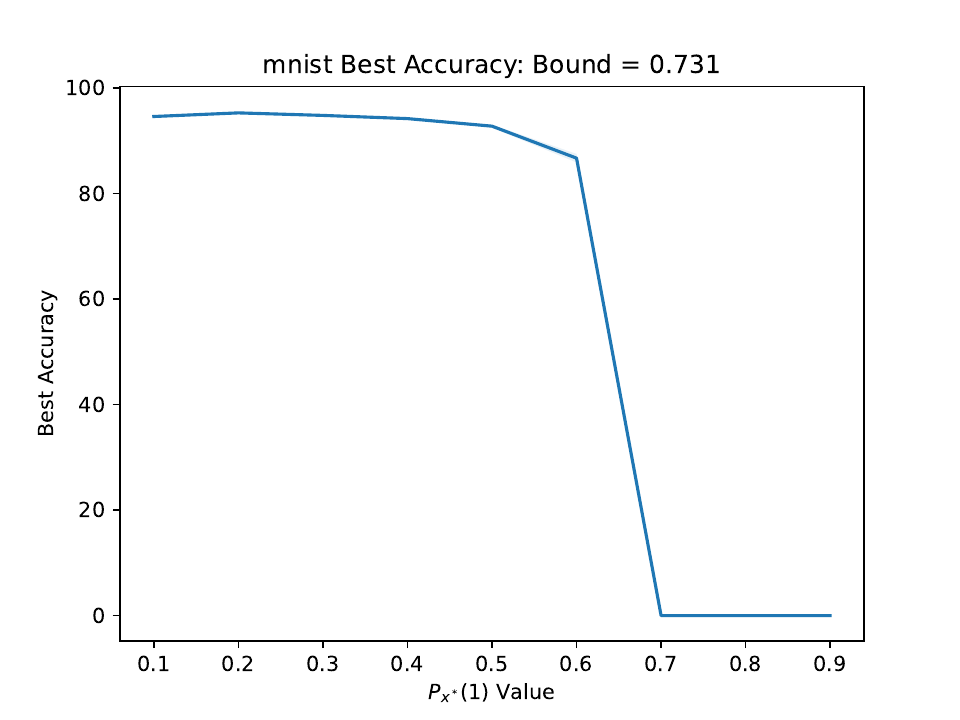}
         \caption{$\varepsilon=1$ when $\mathbb{P}_{\mathbf{x}^*}(1) = 0.5$}
         \label{fig:five over x}
     \end{subfigure}
        \caption{Plots of the best accuracy vs. $\mathbb{P}_{\mathbf{x}^*}(1)$ at differing fixed bounds for MNIST. We include the $95\%$ confidence interval (over the trials).}
        \label{fig:best_acc_mnist}
\end{figure}

\begin{figure}
     \centering
     \begin{subfigure}[b]{0.3\textwidth}
         \centering
         \includegraphics[width=\textwidth]{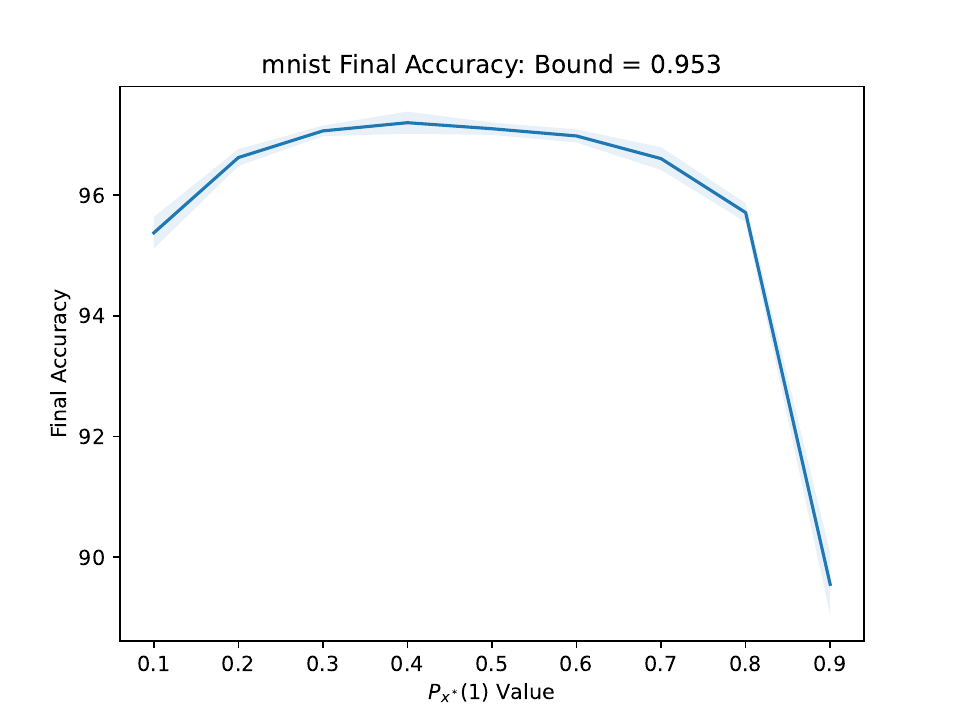}
         \caption{$\varepsilon=3$ when $\mathbb{P}_{\mathbf{x}^*}(1) = 0.5$}
     \end{subfigure}
     \hfill
     \begin{subfigure}[b]{0.3\textwidth}
         \centering
         \includegraphics[width=\textwidth]{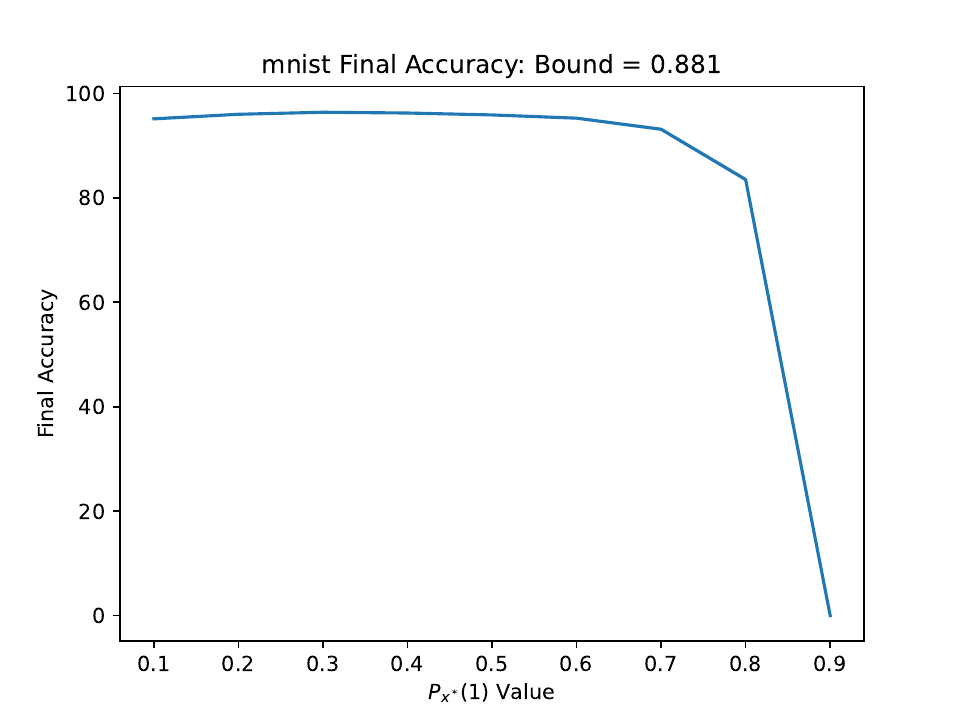}
         \caption{$\varepsilon=2$ when $\mathbb{P}_{\mathbf{x}^*}(1) = 0.5$}
     \end{subfigure}
     \hfill
     \begin{subfigure}[b]{0.3\textwidth}
         \centering
         \includegraphics[width=\textwidth]{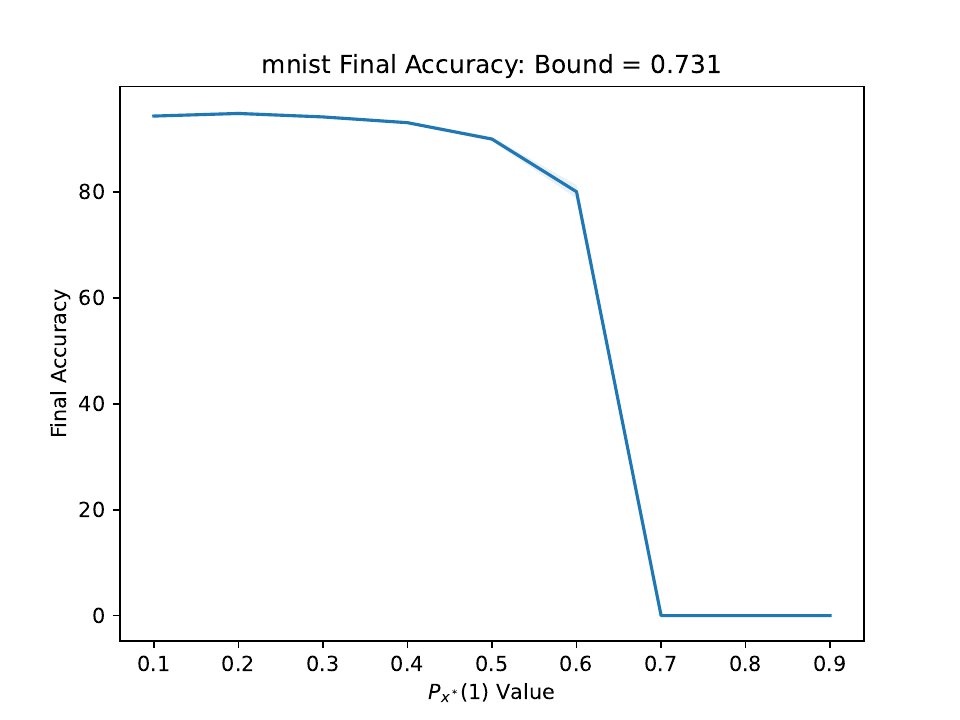}
         \caption{$\varepsilon=1$ when $\mathbb{P}_{\mathbf{x}^*}(1) = 0.5$}
         \label{fig:five over x}
     \end{subfigure}
        \caption{Plots of the final accuracy vs. $\mathbb{P}_{\mathbf{x}^*}(1)$ at differing fixed bounds for MNIST. We include the $95\%$ confidence interval (over the trials).}
        \label{fig:final_acc_mnist}
\end{figure}

To understand the benefits of low sampling (which allows higher $\varepsilon$) versus its effect of smaller training sets for common datasets, we plotted accuracy vs. $P_{\mathbf{x}^*}(1)$ for CIFAR10 and MNIST with different fixed target bounds derived from Theorem~\ref{thm: eps_delta_bound} (setting $\delta =10^{-5}$ always, and considering at most $10\%$ failure, \ie~$P(S|\mathbf{x}^*) = 0.1$). Note the fixed target bounds were chosen to be those achieved when setting $P_{\mathbf{x}^*}(1) = 0.5$ and taking $\varepsilon = 3,2,1$. The models and training hyperparameters (learning rate, mini-batch size) are those used by \cite{tramer2020differentially}, with the hyperparameters being those reported to perform the best for $\varepsilon = 3$. The results (plotting the average over 5 trials) are shown in~\Cref{fig:best_acc_cifar10,fig:final_acc_cifar10,fig:best_acc_mnist,fig:final_acc_mnist}, where note final accuracy is the accuracy after training for $100$ epochs and best accuracy is the highest achieved accuracy during the training run. We observe that MNIST often finds an optimal accuracy with $\mathbb{P}_{\mathbf{x}^*}(1) =0.5$. CIFAR10 suffers somewhat more disproptionately from having smaller epsilons than having smaller datasets, tending to have an optimal accuracy with $\mathbb{P}_{\mathbf{x}^*}(1) = 0.3$. This is even more apparent when looking at the final accuracy for CIFAR10, where $\mathbb{P}_{\mathbf{x}^*}(1) = 0.1$ often performs best. We conclude that CIFAR10 performance \emph{benefits from low sampling (with some fixed target MI bound)}, though further work is needed to understand the benefits for other harder/larger datasets.

Nevertheless, we believe our results illustrate a general point of consideration when implementing DP for common DNNs: to balance the trade-off between dataset size and $\varepsilon$.

\section{Importance to Data Deletion}
\label{ssec:data_deletion}

The ability to decrease MI accuracy, \ie the ability for an arbitrator to attribute a data point to a model, has consequences for machine unlearning and data deletion. %
\citet{cao2015towards} were the first to describe a setting where it is important for the model to be able to "forget" certain training  points. The authors focused on the cases where there exist efficient analytic solutions to this problem. The topic of machine unlearning was then extended to DNNs by \citet{bourtoule2019machine} with the definition that a model has unlearned a data point if after the unlearning, the distribution of models returned is identical to the one that would result from not training with the data point at all. This definition was also stated earlier by \citet{ginart2019making} for other classes of machine learning models.

Given that unlearning is interested in removing the impact a data point had on the model, further work employed MI accuracy on the data point to be unlearned as a metric for how well the model had unlearned it after using some proposed unlearning method~\citep{baumhauer2020machine,graves2020amnesiac,golatkar2020forgetting,golatkar2020eternal}. Yet, empirical estimates on the membership status of a datapoint are subjective to the concrete MI attacks employed. Indeed it may be possible that there exists a stronger attack. The issue of empirical MI for measuring unlearning was recently illustrated by \cite{thudi2021unrolling} where they highlighted undesirable disconnects between it and other measures of unlearning.

\paragraph{Applying our result to unlearning.} Analytic bounds to MI attacks, on the other hand, resolve the subjectivity issue of MI as a metric for unlearning as they bound the success of any adversary. In particular one could give the following definition of an unlearning guarantee from a formal MI positive accuracy bound:

\begin{definition}[$B$-MI Unlearning Guarantee]
An algorithm is $B$-MI unlearnt for $\mathbf{x}^*$ if $\mathbb{P}(\hat{\mathbf{x}}^* \notin D|S) \geq B$, \ie the probability of $\mathbf{x}^*$ not being in the training dataset is greater than $B$.
\end{definition}

Therefore, our result bounding positive MI accuracy has direct consequences on the field of machine unlearning. %
In particular if $\mathbb{P}(\mathbf{x}^* \in D|S)$ is sufficiently low, that is the likelihood of $S$ coming from $\mathbf{x}^*$ is low, then an entity could claim that they do not need to delete the users data since their model is most likely independent of that data point as it most likely came from a model without it: \ie leading to plausible deniability. Note we defined this type of unlearning 
as a $B$-MI unlearning guarantee. This is similar to the logic presented by~\citet{sekhari2021remember} where unlearning is presented probablistically in terms of $(\varepsilon,\delta)$-unlearning.

We also observe an analogous result to~\citet{sekhari2021remember} where we can only handle a maximum number of deletion requests before no longer having sufficiently low probability. To be exact, let us say we do not need to undergo any unlearning process given a set of data deletion request $\hat{\mathbf{x}}^*$ if $\mathbb{P}(\hat{\mathbf{x}}^* \notin D|S) \geq B$ for some $B$ (\ie we are working with probability of not having that set of data in our training set which we want to be high). Note that we sampled data independently, thus if $\hat{\mathbf{x}}^* = \{\mathbf{x}_1^*,\mathbf{x}_2^*,\cdots,\mathbf{x}_m^*\}$, then $\mathbb{P}(\hat{\mathbf{x}}^* \notin D|S) = \mathbb{P}(\mathbf{x}_1^* \notin D|S) \cdots \mathbb{P}(\mathbf{x}_m^* \notin D|S)$.

Now, for simplicity, assume the probability of drawing all points into the datasets are the same, so that for all $\mathbf{x}_i^*$ we have the same bound given by Corollary \ref{cor:MI_neg_acc}, that is $\mathbb{P}(\mathbf{x}_i^* \notin D|S) > L$ for some $L \leq 1$. Then we have $\mathbb{P}(\hat{\mathbf{x}}^* \notin D|S) \geq L^m$ and so an entity does not need to unlearn if $L^m \geq B$, \ie if $m \leq \frac{\ln{1/B}}{\ln{1/L}} = \frac{\ln{B}}{\ln{L}}$. This gives a bound on how many deletion requests the entity can avoid in terms of the lower bound given in Corollary~\ref{cor:MI_neg_acc}. In particular, note that if $\{\mathbf{x}_1 \cdots \mathbf{x}_N\}$ is the larger set of data points an entity is sampling from, and $\mathbb{P}_{\mathbf{x}}(1) = c/N$ $\forall \mathbf{x} \in \{\mathbf{x}_1 \cdots \mathbf{x}_N\}$, then the lower bound given by Corollary~\ref{cor:MI_neg_acc} is $\left({1+\frac{e^{-\varepsilon}\frac{c}{N}}{1-\frac{c}{N}}}\right)^{-1}$.

\begin{corollary}[$B$-MI Unlearning Capacity]
\label{cor:unl_bound}
If $\{\mathbf{x}_1 \cdots \mathbf{x}_N\}$ is the larger set of data points an entity is sampling from, and $\mathbb{P}_{\mathbf{x}}(1) = c/N$ $\forall \mathbf{x} \in \{\mathbf{x}_1 \cdots \mathbf{x}_N\}$, then one can delete $m \leq \frac{\ln{B}}{\ln{L(c)}}$ with $L(c) = \left({1+\frac{e^{-\varepsilon}\frac{c}{N}}{1-\frac{c}{N}}}\right)^{-1}$ and satisfy $B$-MI unlearning.
\end{corollary}

\begin{figure}[h]
    \centering
    \includegraphics[width=70mm]{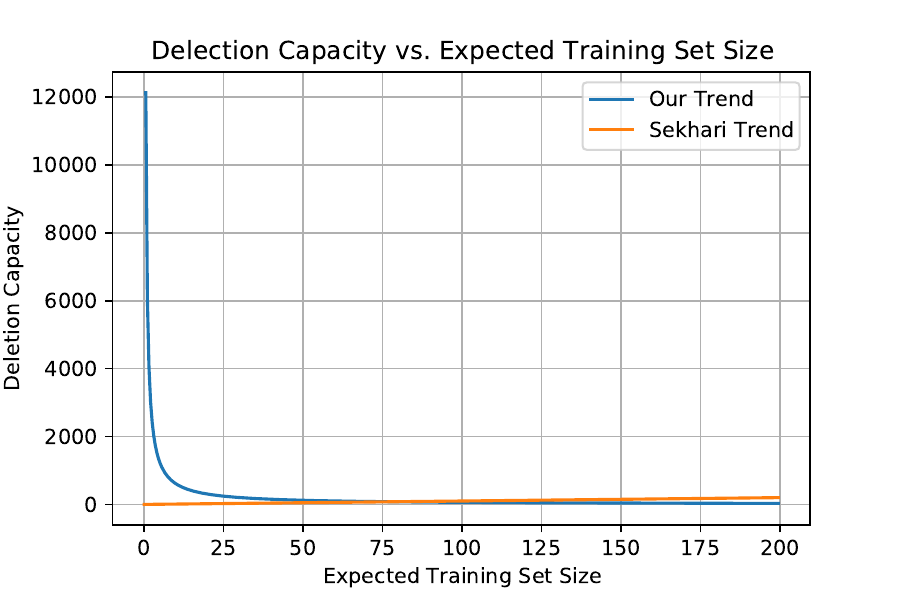}
    \caption{Comparing our deletion capacity trend to the trend~\cite{sekhari2021remember} describes. In particular our number of deletions degrades with training size while theirs increasing.}
    \label{fig:del_capacity}
\end{figure}

\citet{sekhari2021remember} showed that with typical DP the deletion requests grow linearly with the size of the training (in the above case $c$ represents the expected training set size). We thus compare a linear line w.r.t to $c$ to $\frac{\ln{B}}{\ln{L(c)}}$ (given by Corollary~\ref{cor:unl_bound}) in~\Cref{fig:del_capacity} to observe their respective magnitude: we fix $B=0.8$, $N = 10000$ and $\varepsilon = 1$ as we are interested in asymptotics. We observe that our deletion capacity is significantly higher for low expected training dataset sizes and is marginally lower than a linear trend for larger training set sizes.

\section{Discussion}
\label{sec:discussion}

We break our discussion into two themes. The first concerns discussion on how bounds on MI precision can be directly applied to bound other attacks/properties of interest; the objective is to understand the utility of only studying MI attacks. In \S~\ref{ssec:group_inference} we discuss how our bounds can be applied to bound "group" membership inference. In \S~\ref{ssec:reconstruction} we discuss how membership inference precision bounds limit the confidence any reconstruction attack can have, and point out an alternative question of if MI precision bounds can lead to lower-bounds on the error of "interesting" reconstruction attacks. Lastly, in \S~\ref{ssec:generalization} we remark that MI accuracy bounds lead to bounds on the generalization gap of machine learning models, and point out this approach to bounding properties of interest could be generally applicable.

The second half of the discussion focuses on our work in relation to past work.

\subsection{Generalizing to Group Membership Inference}
\label{ssec:group_inference}

One could consider a more general membership inference problem, where instead of asking if we trained on a particular datapoint $\mathbf{x}^*$, we ask if we trained on any $\mathbf{x}^* \in G$ for some set $G \subset \mathbb{D} \coloneqq \{\mathbf{x}_1,\cdots, \mathbf{x}_N\}$. For example, this could be relevant when the set $G$ defines a subgroup: \eg~all people born in a particular city. Hence, inference reveals how much a final model can leak about these specific group features. Bounding the precision of this group membership is the same as bounding $\mathbb{P}(G \cap D \neq \emptyset|S)$. We can obtain a bound on this as a corollary of Theorem~\ref{thm:MI_pos_acc}; the crux of the proof is noting our bounds are invariant to the other datapoints in the larger set of possible datapoints, and this gives us a type of independence at the level of applying our bounds.

\begin{corollary}[Group Membership Inference]
\label{cor:group_MI}

Consider a subset of the set of possible data $G \subset \mathbb{D}$, and let $D$ be the training set obtained following $\varepsilon$-DP setting we have considered so far in the paper, described in Section~\ref{ssec:scenario}. We then have $\mathbb{P}(G \cap D \neq \emptyset|S) \leq 1 - \prod_{\mathbf{x}^* \in G} \left({1+\frac{e^{\varepsilon}\mathbb{P}_{\mathbf{x}^*}(1)}{\mathbb{P}_{\mathbf{x}^*}(0)}}\right)^{-1}$.

\end{corollary}

\begin{proof}
Let $G = \{\mathbf{x}_{i_1},\cdots, \mathbf{x}_{i_g}\}$ for $g = |G|$. Note $\mathbb{P}(G \cap D \neq \emptyset|S) = 1 - \mathbb{P}(\mathbf{x}_{i_j} \not \in D, \forall j \in [g]| S) $. Now note $\mathbb{P}(\mathbf{x}_{i_j} \not \in D, \forall j \in [g]| S) = \prod_{j \in [g]} \mathbb{P}(\mathbf{x}_{i_j} \not \in D| S, \mathbf{x}_{i_{j'}} \not \in D~for~j' < g)$. Note the condition $\mathbf{x}_{i_{j'}} \not \in D~for~j' < g$ restricts us to considering $\mathbb{D} \coloneqq \mathbb{D} \setminus \{\mathbf{x}_{i_{j'}}~for~j' < g\}$, but our bounds in Corollary~\ref{cor:MI_neg_acc} did not depend on what $\mathbb{D}$ exactly were. So we can apply the lower-bound from Corollary~\ref{cor:MI_neg_acc} for each term in the product, and hence get the upper-bound $\mathbb{P}(G \cap D \neq \emptyset|S) \leq 1 - \prod_{\mathbf{x}^* \in G} \left({1+\frac{e^{\varepsilon}\mathbb{P}_{\mathbf{x}^*}(1)}{\mathbb{P}_{\mathbf{x}^*}(0)}}\right)^{-1}$ as desired. 

\end{proof}

Note an analogous bound can be obtained for $(\varepsilon, \delta)$-DP by using Theorem~\ref{thm: eps_delta_bound} to get a lower-bound on $\mathbb{P}(\mathbf{X}^* \not \in D|S)$, which is what we needed for the proof.

\subsection{Applying to Data Reconstruction}
\label{ssec:reconstruction}

An interesting question is how training with DP impacts the effectiveness of data reconstruction attacks. The goal of a data reconstruction attack is, given a model (and/or the sequence of models used to obtain it), to have an algorithm that outputs a training point: more realistically, a point close to a training point. In this sense, the attack "reconstructed" a training datapoint. An immediate consequence of our MI positive accuracy bounds is a statement that data reconstruction attacks are not trustworthy; even when they reconstruct an image close to a possible datapoint, the probability of the datapoint actually being a training datapoint is bounded.

\begin{corollary}[Data Reconstruction not Trustworthy]
Consider a deterministic \footnote{One could consider random algorithms, but fixing the random seed the bound will apply. As the bound applies for every random seed, it also applies in expectation} "data reconstruction" algorithm $R$ that takes a model (or the training sequence of models) $M$ and outputs a datapoint. We say it "reconstructs" a training datapoint out of the set of possible datapoints $\mathbb{D} \coloneqq \{\mathbf{x}_1,\cdots,\mathbf{x}_N\}$ if for some metric $d$, $d(R(M),\mathbf{x}_i) \leq \varepsilon$ for some $\varepsilon \geq 0$. For any $\mathbf{x}_i$, consider the set $S \coloneqq \{M : d(R(M),\mathbf{x}_i) \leq \varepsilon\}$ of models it claims to have trained on $\mathbf{x}_i$. By Theorem~\ref{thm: eps_delta_bound}, when training with $(\varepsilon,\delta)$-DP, the probability $\mathbf{x}_i$ was actually used to train models in $S$ is $ \mathbb{P}(\mathbf{x}^* | S) \leq (1 + \frac{e^{-\varepsilon}\mathbb{P}_{\mathbf{x}^*}(0)}{\mathbb{P}_{\mathbf{x}^*}(1)} - \frac{\delta e^{-\varepsilon}\mathbb{P}_{\mathbf{x}^*}(0)}{\mathbb{P}(S | \mathbf{x}^*)})^{-1}$.
\end{corollary}

\begin{proof}
Just an application of Theorem~\ref{thm: eps_delta_bound}
\end{proof}

However, we believe a more interesting question is if, considering a specific class of reconstruction algorithms $R$, we can lower bound the reconstruction error (with high probability). We believe this should be possible, but currently are not sure what an appropriate class of $R$ is. A naive idea is that $d(R(M),\mathbf{x}) \leq \varepsilon$ iff $\mathbf{x} \in D$, that is, the algorithm can only reconstruct datapoints used in the training set. However, the above corollary states no such attacks can exist when training with $\varepsilon$-DP. So it stands this characterization might be too powerful. We leave finding fruitful classes of reconstruction attacks to study for future work.

\subsection{We can Bound Generalization Error}
\label{ssec:generalization}

\citet{yeom2018privacy} studied MI adversaries in the $\mathbb{P}_{\mathbf{x}^*}(1) = 0.5$ setting that took advantage of a generalization gap $R_g$ between training datapoints and non-training datapoints, and showed an adversary whose accuracy is $(R_g/B +1)/2$ (Theorem 2 in \citet{yeom2018privacy}). Note this accuracy is upper and lower-bounded by Theorem \ref{thm:MI_pos_acc} as the theorem bounds the accuracy of all adversaries when $\mathbb{P}_{\mathbf{x}^*}(1) = 0.5$. Thus we have the following corollary.

\begin{corollary}[MI bounds Generalization Error]
For loss function upper-bounded by $B$ and $\varepsilon$-DP training algorithm, we have $(1+e^{\varepsilon})^{-1} \leq (R_g/B +1)/2 \leq (1+e^{-\varepsilon})^{-1}$ where $R_g$ is the \textit{generalization gap} (Definition 3 in \citet{yeom2018privacy}, where their $D$ is our larger set, and their $S$ is the specific training set sampled from the larger set).
\end{corollary}

This tells us how well the $\mathbb{P}_{\mathbf{x}^*}(1) = 0.5$ sampled dataset generalizes to the entire larger set. We believe this general procedure of using specific adversaries with accuracies dependent on values of interest (\eg the generalization gap here) and applying more general bounds to get a range for those values of interest is something future work can expand on.

\subsection{Our Bounds \new{Match and Improve} Earlier Results}

\begin{figure}
    \centering
    \includegraphics[width=65mm]{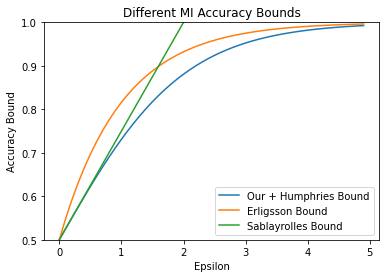}
    \caption{Comparing the upper bound to MI performance we achieved \new{matching \citet{humphries2020differentially}} to that given by~\cite{erlingsson2019we} and~\cite{sablayrolles2019white} (note $\mathbb{P}_{\mathbf{x}^*}(1) = 0.5$ here). In particular note we are tighter for all $\varepsilon$.}
    \label{fig:MI_bounds}
\end{figure}

Note that the setting described in Section \ref{ssec:scenario} is equivalent to the MI experiment defined by \citet{yeom2018privacy} when $\mathbb{P}_{\mathbf{x}_i}(1) = 0.5 ~\forall \mathbf{x}_i \in \{\mathbf{x}_1,\cdots,\mathbf{x}_N\}$. \new{Morover as a consequence of Corollary~\ref{cor:MI_neg_acc}, we have the positive accuracy bounds in Theorem~\ref{thm:MI_pos_acc} also bound MI accuracy for $\mathbb{P}_{\mathbf{x}}(1) = 0.5$. Note the bound we obtain is $(1 + e^{-\varepsilon})$ which matches the bound given by \citet{humphries2020differentially}, the previous tightest bound (see \Cref{fig:MI_bounds}). One of the contributions of our proof is it explicitly demonstrates where the looseness in this bound comes from, \ie~looseness of the DP inequality over all $(D,D’)$ pairs.}

\begin{figure}[h]
    \centering
    \includegraphics[width=70mm]{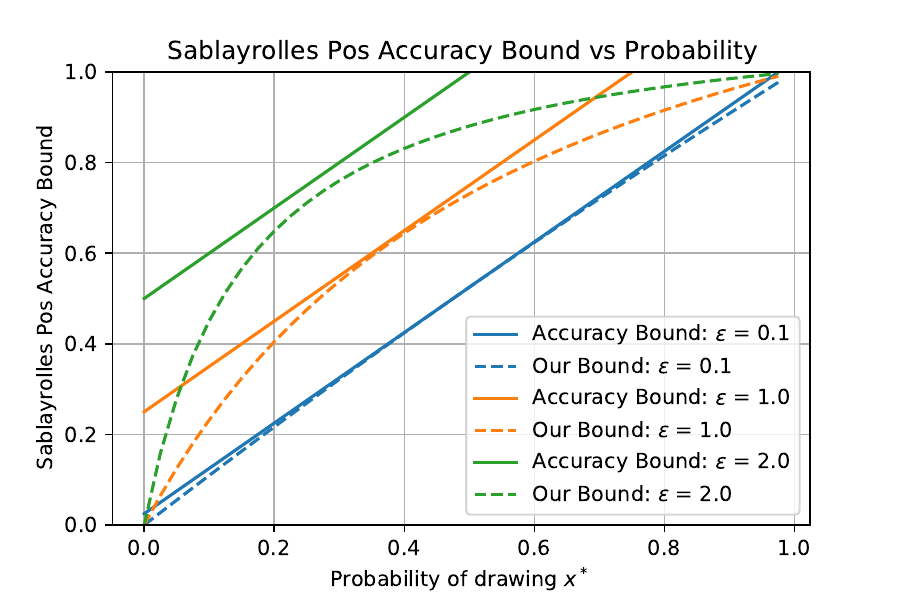}
    \caption{\cite{sablayrolles2019white} upper bound on MI positive accuracy as a function of $\mathbb{P}_{\mathbf{x}^*}(1)$ compared to our bound. Note that we are still tighter for all probabilities.}
    \label{fig:sab_MI_bound_prob}
\end{figure}

\new{Do note our positive accuracy bounds extend to the setting with a biased sampler. We compare our bound with the positive accuracy bound derived by \citet{sablayrolles2019white} in  \Cref{fig:sab_MI_bound_prob} which demonstrates that our bound is indeed tighter.} A reference for past bounds is given in Table~\ref{tb:bounds_table}.

\begin{table}[h]
\begin{center}
\begin{tabular}{|c|c|c|}%
 \textbf{Paper} & \textbf{Analytic Form} & \textbf{Type} \\ 
 \cite{yeom2018privacy} & $e^{\epsilon}/2$ & Accuracy \\  
\cite{erlingsson2019we} & $1-e^{-\epsilon}/2$ & Accuracy \\
\cite{humphries2020differentially} & $(1 + e^{-\varepsilon})$ & Accuracy \\
\cite{sablayrolles2019white} & $\mathbb{P}_{\mathbf{x}^*}(1) + \epsilon/4$ & Positive Accuracy \\
Our Work & $\left({1+\frac{e^{-\epsilon}\mathbb{P}_{\mathbf{x}^*}(0)}{\mathbb{P}_{\mathbf{x}^*}(1)}}\right)^{-1}$ & Positive Accuracy \\
\end{tabular}
\caption{Bounds found in prior work.}
\label{tb:bounds_table}
\end{center}
\end{table}

\subsection{MI Accuracy vs. MI Advantage}

Previous work, particularly~\citet{yeom2018privacy} and~\citet{erlingsson2019we}, focused on membership advantage, which is essentially an improvement in accuracy of \new{an adversary $f$ (with associated set $S$)} over the random guess of $50\%$ of drawing a point in the training dataset. %
More specifically, if we let $A(f)$ denote the accuracy of $f$, then membership advantage is computed as $2(A(f)-0.5)$.
We can generalize this to ask what the membership advantage of the positive accuracy of $f$ compared to the baseline $\mathbb{P}_{\mathbf{x}^*}(1)$ is for values other than only $0.5$.

\begin{figure}
    \centering
    \includegraphics[width=70mm]{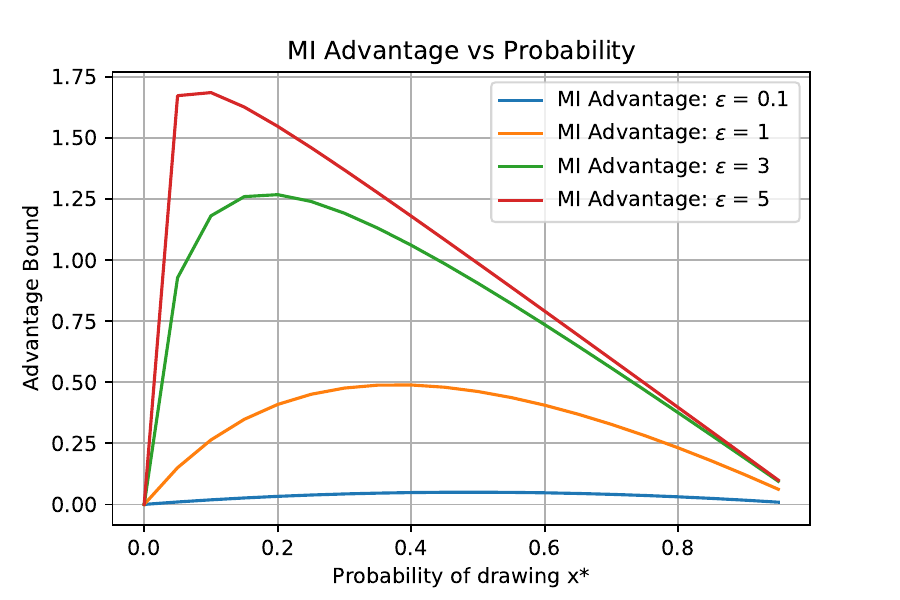}
    \caption{MI advantage}
    \label{fig:MI_adv}
\end{figure}

Theorem \ref{thm:MI_pos_acc} gives us an upper bound on the positive accuracy and thus an upper bound on the positive advantage of $f$ defined as $Ad(f) = 2\left(\mathbb{P}(\mathbf{x}^*|S) - \mathbb{P}_{\mathbf{x}^*}(1)\right) \\ 
\leq 2 \left(\left({1+\frac{e^{-\varepsilon}\mathbb{P}_{\mathbf{x}^*}(0)}{\mathbb{P}_{\mathbf{x}^*}(1)}}\right)^{-1} - \mathbb{P}_{\mathbf{x}^*}(1)\right).$ We plotted this advantage as a function of $\mathbb{P}_{\mathbf{x}^*}(1)$ for different fixed $\varepsilon$ in Figure \ref{fig:MI_adv}. We observe that the advantage clearly depends on $\mathbb{P}_{\mathbf{x}^*}(1)$, and in fact for different $\varepsilon$, the $\mathbb{P}_{\mathbf{x}^*}(1)$ resulting in the maximum advantage changes. In particular, $\mathbb{P}_{\mathbf{x}^*}(1) = 0.5$ is not close to the maximum advantage for large $\varepsilon$, which shows how the fixed experiment proposed by \citet{yeom2018privacy} does not necessarily give the maximum advantage an adversary could have, whereas our result allows us to.

However, it should be noted that higher advantage here does not mean a higher upper bound on MI accuracy; as we already saw in \Cref{fig:MI_bound_prob}, the upper bound on accuracy increases monotonically with $\mathbb{P}_{\mathbf{x}^*}(1)$, in contrast to the bump observed with membership advantage.
This serves to differentiate the study of MI advantage and the study of MI accuracy for future work. 

\section{Conclusion}

In this work, we provide a tight (upto the looseness in the DP guarantee across different datasets) bound on MI positive accuracy against ML models trained with $\varepsilon$-DP, and bound MI positive accuracy when training with $(\varepsilon,\delta)$-DP accepting some necessary probability of failure. Our bound highlights that intricacies of dataset construction are of importance for model MI vulnerability. Indeed, based on our findings, we develop a privacy amplification scheme that just requires one to sub-sample their training dataset from a larger pool of possible data points. While the benefits of such a mechanism for performance are clear for entities with far more data than they need to train on, we empirically illustrated performative benefits for CIFAR10 (having also evaluated on MNIST).
 
Based on our results, entities training their ML models with DP can employ looser privacy guarantees (and thereby preserve their models' accuracy better) while still limiting the success of MI attacks.
Finally, our bound, and more generally bounds on positive MI accuracy, can also be applied to characterize a model's ability to process unlearning requests when defining machine unlearning as achieving a model with low probability of having been derived from a particular data point to be unlearned.

\section*{Acknowledgements}
We would like to acknowledge our sponsors, who support our research with financial and in-kind contributions: Alfred P. Sloan Foundation, Amazon, Apple, Canada Foundation for Innovation, CIFAR through the Canada CIFAR AI Chair program, DARPA through the GARD program, Intel, Meta, NFRF through an Exploration grant, and NSERC through the Discovery Grant and COHESA Strategic Alliance, and the Ontario Early Researcher Award. Resources used in preparing this research were provided, in part, by the Province of Ontario, the Government of Canada through CIFAR, and companies sponsoring the Vector Institute. We would like to thank members of the CleverHans Lab for their feedback.

\bibliographystyle{ACM-Reference-Format}
\bibliography{bibliography}

\appendix

\section{Broader Impact}

Our observation of an independent benefit beyond DP from sampling datapoints with low probability could potentially allow entities with access to significant amounts of data, significantly more than one can reasonably train on, to claim privacy against membership inference by default for high $\epsilon$ when training with $\epsilon$-DP (if they sample i.i.d in some part of the data selection pipeline). However, practically most entities use $(\varepsilon,\delta)$-DP for privacy, and as noted this already comes with a probability of failure for the bound. This probability of failure should be clearly explained to the entity contributing training data. Nevertheless, our work could, in the case where privacy is primarily about membership inference, allow for looser DP parameters avoiding the degradation to performance typically seen with DP. However we emphasize this is only for membership inference, and there are other aspects to privacy beyond membership inference.
\end{document}